\documentclass[10pt,journal,compsoc]{IEEEtran}

\usepackage{url}
\usepackage{ragged2e}
\usepackage{epsfig}
\usepackage{graphicx}
\usepackage{amsmath,amssymb} 
\usepackage{subfigure}
\usepackage{algorithm}
\usepackage{algorithmicx}
\usepackage{algpseudocode}
\usepackage{graphics}
\usepackage{threeparttable}
\usepackage{color}
\usepackage[normalem]{ulem}
\usepackage{multirow}
\usepackage{float}
\usepackage{amsfonts}
\usepackage{bm}
\usepackage{array}
\usepackage[table]{xcolor}
\usepackage{colortbl}
\usepackage{pifont}
\usepackage{diagbox}
\usepackage{rotating}
\usepackage{booktabs}
\usepackage{overpic}
\usepackage{textcomp}
\usepackage{contour}
\usepackage{enumitem}
\usepackage[pagebackref=false,breaklinks=false,colorlinks=false]{hyperref}

\newcommand{\figref}[1]{Fig.~\ref{#1}}
\newcommand{\tabref}[1]{Table~\ref{#1}}

\newcommand{\secref}[1]{$\S$ \ref{#1}}

\RequirePackage{silence}
\graphicspath{{./Imgs/}}
\DeclareGraphicsExtensions{.jpg,.pdf,.png}

\usepackage[labelsep= period]{caption}
\newcommand{\tabincell}[2]{
\begin{tabular}{@{}#1@{}}#2\end{tabular}
}

\def\ie{\emph{i.e.}}
\def\eg{\emph{e.g.}}
\def\etc{\emph{etc}}
\def\etal{{\em et al.~}}
\def\ourmodel{BASNet}

\begin{document}
%
\title{Boundary-Aware Segmentation Network for Mobile and Web Applications}
%
%
%
%

\author{Xuebin Qin,
        Deng-Ping Fan,
        Chenyang Huang,
        Cyril Diagne,
        Zichen Zhang,\\
        Adrià Cabeza Sant'Anna,
        Albert Suàrez,
        Martin Jagersand,
        and Ling Shao,~\IEEEmembership{Fellow,~IEEE}
\IEEEcompsocitemizethanks{\IEEEcompsocthanksitem Xuebin Qin is with the Department
of Computing Science, University of Alberta, Edmonton,
AB, Canada, T6G 2R3.\protect (Email: xuebin@ualberta.ca)
\IEEEcompsocthanksitem 
Deng-Ping Fan and Ling Shao are with the Inception Institute of Artificial Intelligence (IIAI), Abu Dhabi, UAE.
(Email: dengpfan@gmail.com; ling.shao@ieee.org)
\IEEEcompsocthanksitem Chenyang Huang is with the Department of Computing Science, University of Alberta, Edmonton, AB, Canada. (Email: chuang8@ualberta.ca)
\IEEEcompsocthanksitem Cyril Diagne is with Init ML (Email: cyril@initml.co)
\IEEEcompsocthanksitem Zichen Zhang is with the Department of Computing Science, University of Alberta, Edmonton, AB, Canada. (Email: vincent.zhang@ualberta.ca)
\IEEEcompsocthanksitem Adrià Cabeza Sant'Anna is with the Department of Computer Science, Polytechnic University of Catalonia, BarcelonaTech,  Barcelona, Spain. (Email: adriacabezasantanna@gmail.com)
\IEEEcompsocthanksitem Albert Suàrez is with the Department of Software Engineering, Polytechnic University of Catalonia, BarcelonaTech, Barcelona, Spain. (Email: alsumo95@gmail.com)
\IEEEcompsocthanksitem Martin Jagersand is with the Department of Computing Science, University of Alberta, Edmonton, AB, Canada. (Email: mj7@ualberta.ca)
\IEEEcompsocthanksitem A preliminary version of this work has appeared in CVPR~\cite{qin2019basnet}.
\IEEEcompsocthanksitem
Corresponding author: Deng-Ping Fan.
}
\thanks{Manuscript submitted December 6, 2020; revised xx xx, xx.}}

%
%

\markboth{submitted to IEEE XXXX}%
{Shell \MakeLowercase{\textit{et al.}}: Bare Demo of IEEEtran.cls for Computer Society Journals}
%



\IEEEtitleabstractindextext{%
\begin{abstract}
\justifying
Although deep models have greatly improved the accuracy and robustness of image segmentation, obtaining segmentation results with highly accurate boundaries and fine structures is still a challenging problem. 
In this paper, we propose a simple yet powerful Boundary-Aware Segmentation Network (\textbf{BASNet}), which comprises a predict-refine architecture and a hybrid loss, for highly accurate image segmentation. 
The predict-refine architecture consists of a densely supervised encoder-decoder network and a residual refinement module, which are respectively used to predict and refine a segmentation probability map. 
The hybrid loss is a combination of the binary cross entropy, structural similarity and intersection-over-union losses, which guide the network to learn
three-level (\ie, pixel-, patch- and map- level) hierarchy representations.
We evaluate our BASNet on two reverse tasks including salient object segmentation, camouflaged object segmentation, showing that it achieves very competitive performance with sharp segmentation boundaries.
Importantly, BASNet runs at over 70 fps on a single GPU which benefits many potential real applications. Based on BASNet, we further developed two (close to) commercial applications: \textbf{AR COPY \& PASTE}, in which BASNet is integrated with augmented reality for ``COPYING'' and ``PASTING'' real-world objects, and \textbf{OBJECT CUT}, which is a web-based tool for automatic object background removal. 
Both applications have already drawn huge amount of attention and have important real-world impacts. 
The code and two applications will be publicly available at: \url{https://github.com/NathanUA/BASNet}. 



\end{abstract}

\begin{IEEEkeywords}
Boundary-aware segmentation, predict-refine architecture, salient object, camouflaged object
\end{IEEEkeywords}}

\maketitle

\IEEEdisplaynontitleabstractindextext


%
\IEEEpeerreviewmaketitle

\IEEEraisesectionheading{\section{Introduction}\label{sec:introduction}}

\IEEEPARstart{I}{mage} segmentation has been studied over many decades using conventional methods, and in the past few years using deep learning. 
Several different conventional approaches, such as interactive methods, active contour (level-set methods), graph-theoretic approaches, perceptual grouping methods and so on, have been studied for image segmentation over the past decades. 
Yet automatic methods fail where boundaries are complex. Interactive methods let humans resolve the complex cases. Interactive methods,~\cite{DBLP:journals/ijcv/RussellTMF08, DBLP:journals/cviu/WangHC14, DBLP:conf/cvpr/CastrejonKUF17, qin2018bylabel}, are usually able to produce accurate and robust results, but with significant time costs. 
Active contour \cite{osher1988fronts, RC30ActiveContour, RC3ActiveContour, RC63ActiveContour, RC8ActiveContour, caselles1997geodesic, RC67ActiveContour}, graph-theoretic \cite{wu1993optimal, cox1996ratio, shi2000normalized, sarkar2000supervised, wang2003image} and perceptual grouping \cite{GestaltLaws, elder1993effect, mahamud2003segmentation, amir1998generic, elder2003contour, DBLP:conf/iccv/RenFM05, qin2017bmvc, qin2017iros}, methods require almost no human interventions so that they are faster than interactive methods. However, they are relatively less robust. 

In recent years, to achieve accurate, robust and fast performance, many deep learning models \cite{minaee2020image} have been developed for image segmentation. Semantic image segmentation \cite{long2015fully, huang2017densely} is one of the most popular topics, which aims at labeling every pixel in an image, with one of the several predefined class labels. It has been widely used in many applications, such as scene understanding \cite{lin2014microsoft, zhou2017scene}, autonomous driving \cite{Geiger2013IJRR, Cordts2016Cityscapes}, \etc. The targets in these applications are usually large in size, so most existing methods focus on achieving robustness performance with high regional accuracy. Less attention has been paid to the high spatial accuracy of boundaries and fine structures. However, many other applications, \eg~image segmentation/editing \cite{qin2018accurate,DBLP:journals/ijcv/KadirB01,goferman2012context,qin2018bylabel} and manipulation \cite{DBLP:conf/iccv/Jagersand95,DBLP:conf/wacv/MechrezSZ18}, visual tracking \cite{DBLP:conf/wacv/LeeK18,DBLP:conf/iros/QinHQSDJ17,DBLP:conf/bmvc/QinHZDJ17}, vision guided robot hand manipulation \cite{qin2017iros} and so on, require highly accurate object boundaries and fine structures. 


There are two main challenges in accurate image segmentation:
\textbf{Firstly}, large-scale features play important roles in classifying pixels since they can provide more semantic information compared with local features. However, large-scale features are usually obtained from deep low-resolution feature maps or by large size kernels and the spatial resolution is sacrificed. 
Simple upsampling of the low-resolution feature maps to high resolution is not able to recover the fine structures \cite{long2015fully}. 
Thus, many encoder-decoder architectures \cite{ronneberger2015u} have been developed for segmenting edges or thin structures. 
Their skip connections and gradual upsampling operations play important roles in recovering the high resolution probability maps. 
Additionally, different cascaded or iterative architectures \cite{xu2017deep,wang2018detect,deng2018r3net,DBLP:conf/iccv/WangBZZL17} have been introduced to further improve the segmentation accuracy by gradually refining the coarse predictions, which sometimes leads to complicated network architectures and computational bottleneck. 
\textbf{Secondly}, most of the image segmentation models use cross entropy (CE) loss to supervise the training process. 
CE loss usually gives greater penalties on these seriously erroneous predictions (\eg~predict ``1'' as ``0.1'' or predict ``0'' as ``0.9''). 
Therefore, deep models trained with CE loss compromise andes prefer to predict ``hard'' samples with a non-committed ``0.5''. In image segmentation tasks, the boundary pixels of targets are usually the hard samples, so this will lead to blurry boundaries in predicted segmentation probability maps. 
Other losses, such as intersection-over-union (IoU) loss \cite{rahman2016optimizing,mattyus2017deeproadmapper,DBLP:conf/bmvc/NagendarSBJ18}, F-score loss \cite{DBLP:journals/corr/abs-1805-07567}  and Dice-score loss \cite{DBLP:conf/miccai/FidonLGEKOV17a}, 
have also been introduced to image segmentation tasks for handling biased training sets. 
These are sometimes able to achieve higher (regional) evaluation metrics, \eg~IoU, F-score, since their optimization targets are consistent with these metrics. 
However, they are not specifically designed for capturing fine structures and often produce ``biased'' results, which tend to emphasize the large structures while neglecting fine details. 

\begin{figure}[t!]
	\centering
	\begin{overpic}[width=\columnwidth]{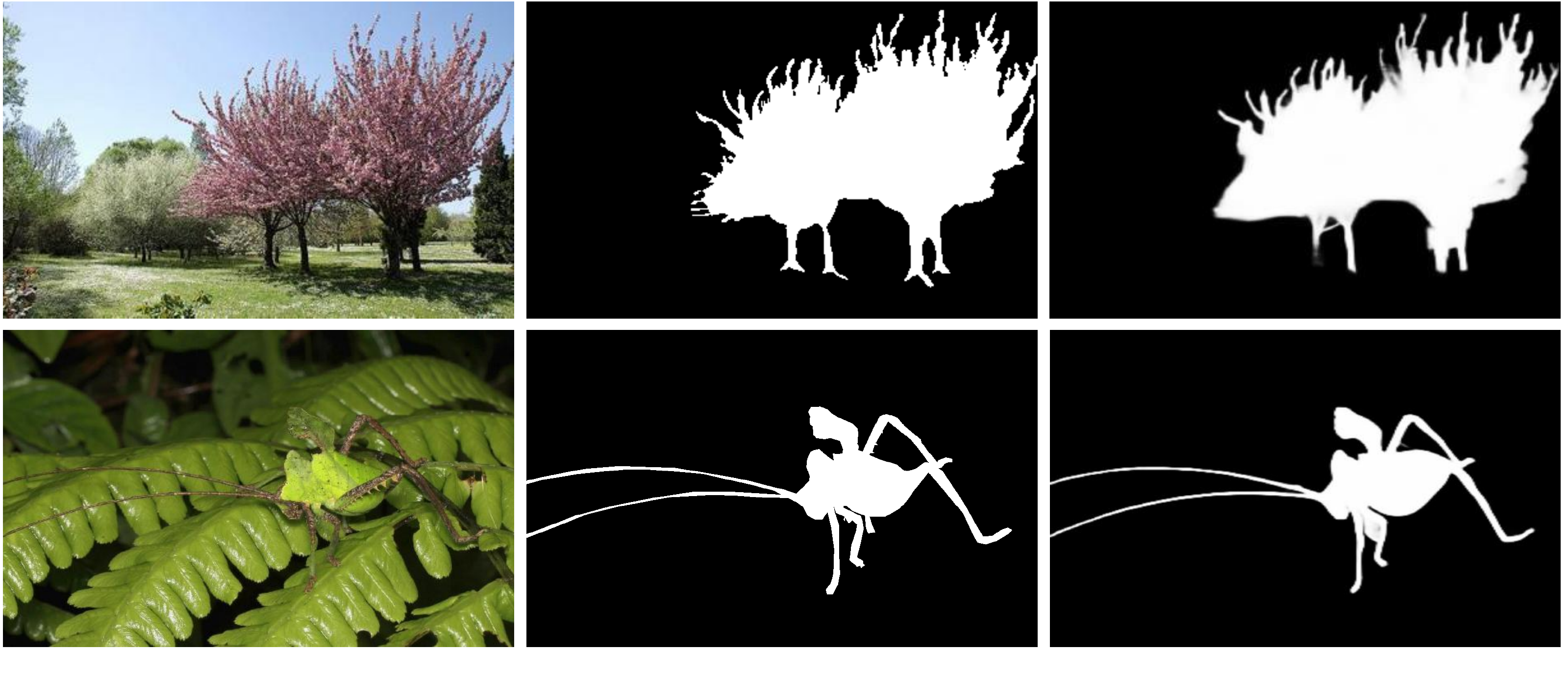}
    \put(7,-3) {(a) Image}
    \put(45,-3) {(b) GT}
    \put(75,-3) {(c) BASNet}
  \end{overpic}
  \vspace{2pt}
  \caption{Sample results of our BASNet on salient object detection (Top) and camouflaged object detection (Bottom).
  }\label{fig:intro}
\end{figure}

To address the above issues, we propose a novel but simple \textbf{B}oundary-\textbf{A}ware \textbf{S}egmentation \textbf{Net}work (\textbf{BASNet}), which consists of a predict-refine network and a hybrid loss, for highly accurate image segmentation. 
The predict-refine architecture is designed to predict and refine the predicted probability maps sequentially. 
It consists of a U-Net-like \cite{ronneberger2015u} deeply supervised \cite{lee2015deeply,xie2015holistically} ``heavy'' encoder-decoder network and a residual refinement module with ``light'' encoder-decoder structure. The ``heavy'' encoder-decoder network transfers the input image to a segmentation probability map, while the ``light'' refinement module refines the predicted map by learning the residuals between the coarse map and ground truth (GT). 
In contrast to  \cite{peng2017large, islamsalient, deng2018r3net}, which iteratively use refinement modules on saliency predictions or intermediate feature maps at multiple scales, our refinement module is used only once on the original scale of the segmentation maps. Overall, our predict-refine architecture is concise and easy to use. 
The hybrid loss combines the binary cross entropy (BCE) \cite{DBLP:journals/anor/BoerKMR05}, structural similarity (SSIM) \cite{wang2003multiscale} and IoU losses \cite{mattyus2017deeproadmapper}, to supervise the training process in a three-level hierarchy: pixel-, patch- and map- level. 
Instead of using \textit{explicit} boundary loss (NLDF+ \cite{luo2017non}, C2S \cite{DBLP:conf/eccv/LiYCLS18}, BANet\cite{su2019selectivity}), 
we \textit{implicitly} inject the goal of accurate boundary prediction in the hybrid loss, contemplating that it may help reduce spurious error from cross propagating the information learned from the boundaries and other regions on the image (see \figref{fig:intro}).

In addition to proposing novel segmentation techniques, developing novel segmentation applications also plays a very important role in advancing the segmentation field. 
Therefore, we developed two novel BASNet-based applications: \textbf{AR COPY \& PASTE}\footnote{\url{https://clipdrop.co/}} and \textbf{OBJECT CUT}\footnote{\url{https://objectcut.com/}}. 
AR COPY \& PASTE is a mobile app built upon our BASNet model and the Augmented Reality techniques. 
By using cellphones, it provides a novel interactive user experience where users can ``COPY'' the real-world targets and ``PASTE'' them into desktop software . 
Specifically, AR COPY \& PASTE allows users to take a photo of an object using a mobile device. Then the background removed object returned by our remote BASNet server will be shown in the camera view. In this view, the ``COPIED“ object is overlapped with real scene video stream. 
Users can move and target the mobile camera at a specific position on the desktop screen. 
Then tapping the screen of the mobile device will trigger the ``PASTE'' operation, which transmits the object from the mobile device to the software opened in the desktop. Meanwhile, OBJECT CUT provides a web-based service for automatic image background removal based on our BASNet model. An image can be uploaded from a local machine or through an URL. This application greatly facilitates the background removal for users who have no image editing experience or software.
The main contributions can be summarized as:
\begin{itemize}[noitemsep]
    \item
    We develop a novel boundary-aware image segmentation network, BASNet, which consists of a deeply supervised encoder-decoder and a residual refinement module, and a novel hybrid loss that fuses BCE, SSIM, and IoU to supervise the training process of accurate image segmentation on three levels: pixel-level, patch-level and map-level.
    \item
    We conduct thorough evaluations of the proposed method including a comparison with 25 state-of-the-art (SOTA) methods on six widely used public salient object segmentation datasets, a comparison with 16 models on the SOC (Salient Object in Clutter) dataset and a comparison with 13 camouflaged object detection (COD) models on three public COD datasets. \ourmodel~achieves very competitive performance in terms of regional evaluation metrics, while outperforms other models in terms of boundary evaluation metrics.
    \item 
    We develop two (close to) commercial applications, AR COPY \& PASTE and OBJECT CUT, based on our BASNet. These two applications further demonstrate the simplicity, effectiveness and efficiency of our model.
\end{itemize}

Compared with the CVPR version~\cite{qin2019basnet} of this work, the following extensions are made. First, deeper theoretical explanations of the hybrid loss design are added. Second, more comprehensive and thorough experiments on different datasets, including salient objects in clutter (SOC) and COD, are included. Third, two (close to) commercial applications, AR COPY \& PASTE and OBJECT CUT, are developed. 

\section{Related Works}
\subsection{Traditional Image Segmentation Approaches} 
Watershed \cite{vincent1991watersheds}, graph cut \cite{kumar2005obj,ladicky2010graph}, active contour \cite{osher1988fronts}, perceptual grouping \cite{qin2017bmvc} as well as the interactive methods based on these approaches mainly rely on well-designed handcrafted features, objective functions and optimization algorithms. 
Watershed and graph cut approaches segment images based on the regional pixel similarities, so they are less effective in segmenting very fine structures and achieving smooth and accurate segmentation boundaries. 
Active contour and perceptual grouping methods can be considered as boundary based approaches. Active contour methods represent the 2D segmentation contour by the level set of a 3D function. Instead of directly evolving the 2D contour, this group of approaches evolves the 3D function to find the optimal segmentation contour, which avoids complicated 2D contour splitting and merging issues. Perceptual grouping methods segment images by selecting and grouping subsets of the detected edge fragments or line segments from given images to formulate closed or open contours of the targets to be segmented. However, although these methods are able to produce relatively accurate boundaries, they are very sensitive to noise and local minima, which usually leads to less robust and unreliable performance. 

\subsection{Patch-wise Deep Models} 
To improve the robustness and accuracy, deep learning methods have been widely introduced to image segmentation \cite{qin2020visual}. Early deep methods use existing image classification networks as feature extractors and formulate the image segmentation tasks as patch-wise image pixel (super-pixel) \cite{li2015visual, liu2015predicting, wang2015deep, zhao2015saliency, li2016visual} classification problems. These models greatly improve the segmentation robustness in some tasks due to the strong fitting capability of deep neural networks. However, they are still not able to produce high spatial accuracy, let alone segmenting fine structures. The main reason is probably that the pixels in patch-wise models are classified independently based on the local features inside each patch and larger-scale spatial contexts are not used.

\subsection{FCN and its Variants} 
With the development of fully convolutional network (FCN) \cite{long2015fully}, deep convolutional neural networks have become a standard solution for image segmentation problems. 
Large number of deep convolutional models \cite{minaee2020image} have been proposed for image segmentation. FCN adapts classification backbones, such as VGG \cite{simonyan2014very}, GoogleNet \cite{szegedy2015going}, ResNet \cite{he2016deep} and DenseNet \cite{huang2017densely}, by discarding the fully connected layers and directly upsampling the output features of certain convolutional layers with specific scales to build a fully convolutional image segmentation model. However, the direct upsampling from low resolution fails in capturing accurate structures. Therefore, The DeepLab family \cite{chen2014semantic, chen2017deeplab, chen2017rethinking} replaces the pooling operations by atrous convolutions to avoid degrading the feature map resolution. Besides, they also introduce a densely connected Conditional Random Field (CRF) to improve the segmentation results. However, applying atrous convolutions on high-resolution maps leads to larger memory costs and CRF usually yields noisy segmentation boundaries. 
Holistically Edge Detection (HED) \cite{xie2015holistically}, RCF \cite{girshick2014rich} and CASENet \cite{yu2017casenet} are proposed to directly segment edges by making full use of the features from both the shallow and deep stages of the image classification backbones. 
Besides, many variants \cite{li2016deep, kruthiventi2016saliency,hu2017deep} of FCN have been proposed for salient object detection (binary-class image segmentation) \cite{wang2019salient}. 
Most of these works are focusing on either developing novel multi-scale feature aggregation strategies or designing new multi-scale feature extraction modules. 
Zhang \etal (Amulet) \cite{DBLP:conf/iccv/ZhangWLWR17} developed a generic framework for aggregating multi-level convolutional features of the VGG backbone. 
Inspired by HED \cite{xie2015holistically}, Hou \etal (DSS+) \cite{hou2017deeply} introduced short connections to the skip-layer structures of HED to better use the deep layer features.  
Chen \etal (RAS) \cite{DBLP:conf/eccv/ChenTWH18} developed a reverse attention model to iteratively refine the side-outputs from a HED-like architecture. 
Zhang \etal (LFR) \cite{DBLP:conf/ijcai/ZhangLLS18} designed a symmetrical fully convolutional network which takes images and their reflection as inputs to learn the saliency features from the complementary input spaces. 
Instead of passing the information with single direction (deep to shallow or shallow to deep), Zhang \etal (BMPM) \cite{zhang2018bi} proposed to have the information passed between the shallow and deep layers by a controlled bi-directional passing module.

\subsection{Encoder-decoder Architectures} 
Rather than directly upsampling features from deep layers of the backbones, SegNet \cite{badrinarayanan2017segnet} and U-Net \cite{ronneberger2015u} employ encoder-decoder like structures to gradually up-sample the deep low-resolution feature maps. Combined with skip connections, they are able to recover more details. 
One of the main characteristics of these models is the symmetrical downsampling and upsampling operations. 
To reduce the checkerboard artifacts in the prediction, Zhang \etal (UCF) \cite{DBLP:conf/iccv/ZhangWLWY17} reformulated the dropout and developed a hybrid module for the upsampling operation. 
To better use the features extracted by backbones, Liu \etal (PoolNet) \cite{liu2019simple} constructed the decoder part using their newly developed feature aggregation, pyramid pooling and global guidance modules. 
In addition, stacked HourglassNet \cite{newell2016stacked}, CU-UNet \cite{tang2018quantized}, UNet++ \cite{zhou2018unet++} and U$^2$-Net \cite{qin2020u2} further explore diverse ways of improving the encoder-decoder architectures by cascaded or nested stacking.


\begin{figure*}[t!]
	\centering
	\begin{overpic}[width=\textwidth]{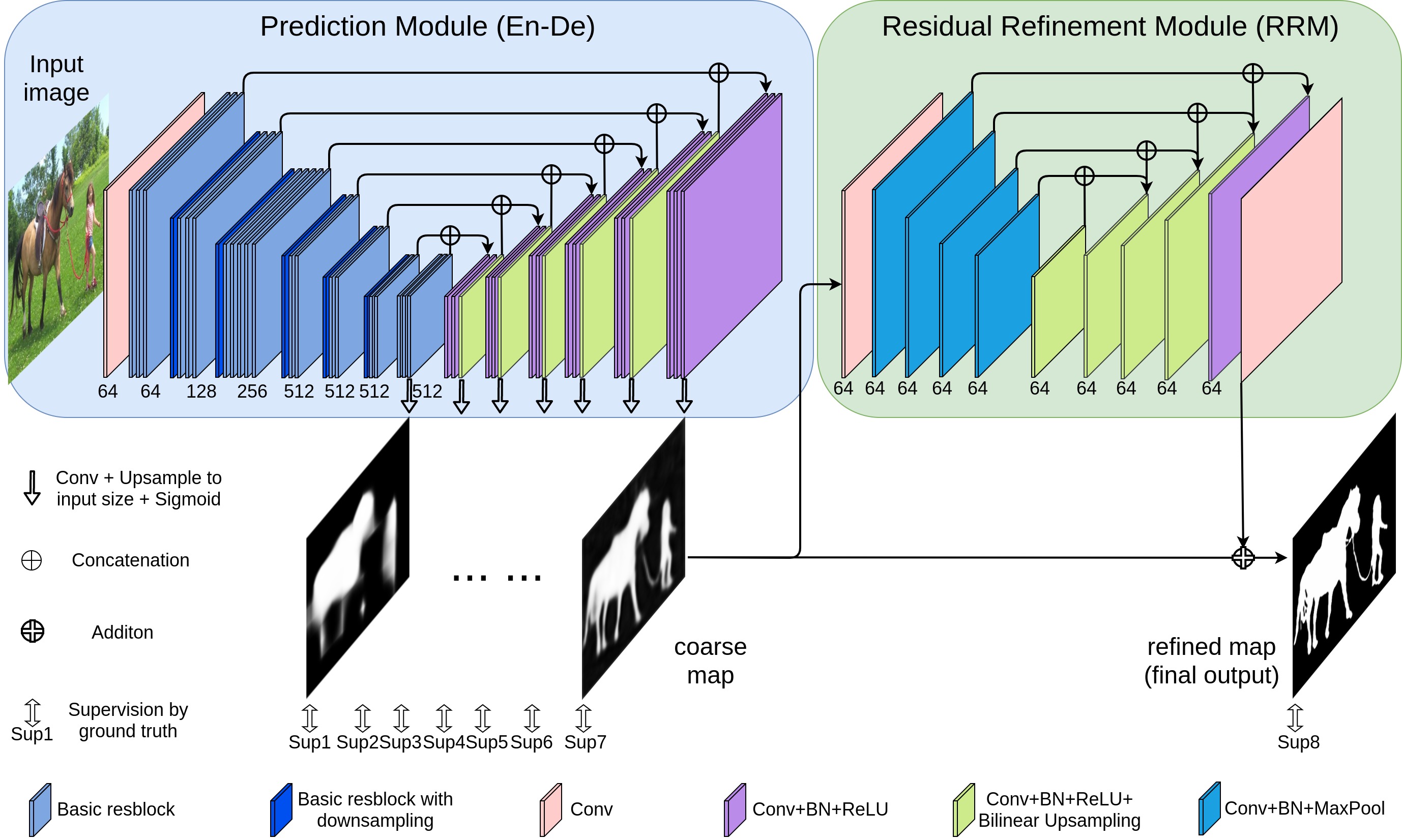}
  \end{overpic}
  \caption{Architecture of the proposed boundary-aware segmentation network: BASNet. See \secref{sec:Methodology} for details.
  }\label{fig:arc}
\end{figure*}

\subsection{Deep Recurrent Models} 
Recurrent techniques have been widely applied in image segmentation models. 
Kuen \etal \cite{kuen2016recurrent} proposed to achieve the completely segmented results by sequentially segmenting image sub-regions using a recurrent framework. 
Liu \etal (PiCANetR) \cite{liu2018picanet} deployed a bidirectional LSTM along the row and column of the feature maps respectively to generate the pixel-wise attention maps for salient object detection. 
Hu \etal (SAC-Net) \cite{hu2020sac} developed similar strategies to~\cite{liu2018picanet} to capture spatial attenuation context for image segmentation. 
Zhang \etal (PAGRN) \cite{zhang2018progressive} proposed to transfers global information from  deep to shallower layers via a multi-path recurrent connection. 
Wang \etal (RFCN) \cite{wang2018salient} built a cascaded network by stacking multiple encoder-decoders to recurrently correcting the prediction errors of the previous stages. 
Instead of iteratively refining the segmentation results \cite{wang2018salient}, Hu \etal (RADF+) \cite{DBLP:conf/aaai/HuZQFH18} recurrently aggregated and refined multi-layer deep features to achieve accurate segmentation results. 
However, due to the serial connections between each recurrent step, models using the ``recurrent'' techniques are relatively less efficient in terms of time costs. 

\subsection{Deep Coarse-to-Fine Models} 
This group of models aims at improving the segmentation results by gradually refining the coarse predictions. 
Lin \etal (RefineNet) \cite{lin2017refinenet} developed a multi-path refinement segmentation network, which uses long-range residual connections to exploit the information along the down-sampling process. 
Liu \etal (DHSNet) \cite{liu2016dhsnet} proposed a hierarchical recurrent
convolutional neural network (HRCNN), which hierarchically and progressively refines the segmentation results in a coarse-to-fine manner. 
Wang \etal (SRM) \cite{DBLP:conf/iccv/WangBZZL17} developed a multi-stage framework for segmentation map refinement, in which each stage takes the input image and the segmentation maps (lower resolution) from the last stage to produce higher-resolution results. 
Deng \etal (R$^3$Net+) \cite{deng2018r3net} proposed to alternatively refine the segmentation results based on the shallow, high-resolution and deep low-resolution feature maps. 
Wang \etal (DGRL) \cite{wang2018detect} developed a global-to-local framework which first localizes the to-be-segmented targets globally and then refines these targets using a local boundary refinement module. 
The coarse to fine models reduce the probability of overfitting and show promising improvements in accuracy. 

\subsection{Boundary-assisted Deep Models} 
Region and boundaries are mutually determined. Therefore, many models introduce boundary information to assist segmentation. 
Luo \etal (NLDF) \cite{luo2017non} proposed to supervise a 4$\times$5 grid structure adapted from VGG-16 by fusing the cross entropy and the boundary IoU inspired by Mumford-Shah \cite{mumford1989optimal}. 
Li \etal (C2S) \cite{DBLP:conf/eccv/LiYCLS18} tried to recover the regional saliency segmentation from segmented contours. 
Su \etal (BANet) \cite{su2019selectivity} developed a boundary-aware segmentation network with three separate streams: a boundary localization stream, an interior perception stream and a transition compensation stream for boundary, region and boundary/region transition prediction, respectively. 
Zhao \etal (EGNet) \cite{zhao2019egnet} proposed an edge guidance network for salient object segmentation by explicitly modeling and fusing complementary region and boundary information. 
Most of models in this category explicitly use boundary information as either an additional supervision loss or a assisting prediction stream for inferring the region segments. 


In this paper, we propose a simple predict-refine architecture which takes advantage of both the encoder-decoder architecture and the coarse-to-fine strategy. Besides, instead of explicitly using boundary loss or additional boundary prediction streams, we design a simple hybrid loss which implicitly describes the dissimilarity between the segmentation prediction and the ground truth at three levels: pixel-, patch- and map-level. The predict-refine architecture together with the hybrid loss provides a simple yet powerful solution for image segmentation and some close to commercial applications.

\section{Methodology}\label{sec:Methodology}
\subsection{Overview}
Our BASNet architecture consists of two modules as shown in Fig. \ref{fig:arc}. 
The prediction module is a U-Net-like densely supervised Encoder-Decoder network \cite{ronneberger2015u}, which learns to predict segmentation probability maps from input images. 
The multi-scale Residual Refinement Module (RRM) refines the resulting map of the prediction module by learning the residuals between the coarse map and the GT.


\subsection{Prediction Module}\label{sec:pm}
Share same spirit of U-Net \cite{ronneberger2015u} and SegNet \cite{badrinarayanan2017segnet}, we design our segmentation prediction module as an encoder-decoder fashion, since this kind of architectures is able to capture high-level global contexts and low-level details at the same time. 
To reduce over-fitting, the last layer of each decoder stage is supervised by the GT, inspired by HED \cite{xie2015holistically} (see Fig. \ref{fig:arc}). 
The encoder has an input convolutional layer and six stages comprised of basic res-blocks. 
The input convolutional layer and the first four stages are adopted from ResNet-34 \cite{he2016deep}. 
The difference is that our input layer has 64 convolutional filters with a size of 3$\times$3 and stride of 1 rather than a size of 7$\times$7 and stride of 2. 
Additionally, there is no pooling operation after the input layer.  
This means that the feature maps before the second stage have the same spatial resolution as the input image. 
This is different from the original ResNet-34, which has a quarter of the resolution in the first feature map.
This adaptation enables the network to obtain higher resolution feature maps in earlier layers, while decreasing the overall receptive fields. 
To achieve the same receptive field as ResNet-34 \cite{he2016deep}, we add two more stages after the fourth stage of ResNet-34. 
Both stages consist of three basic res-blocks with 512 filters after a non-overlapping max pooling layer of size 2. 
To further capture global information, we add a bridge stage between the encoder and decoder. 
It consists of three convolutional layers with 512 dilated (dilation=2)~\cite{yu2015multi} 3$\times$3 filters. 
Each of these convolutional layers is followed by a batch normalization \cite{ioffe2015batch} and a ReLU activation function \cite{hahnloser2001permitted}.

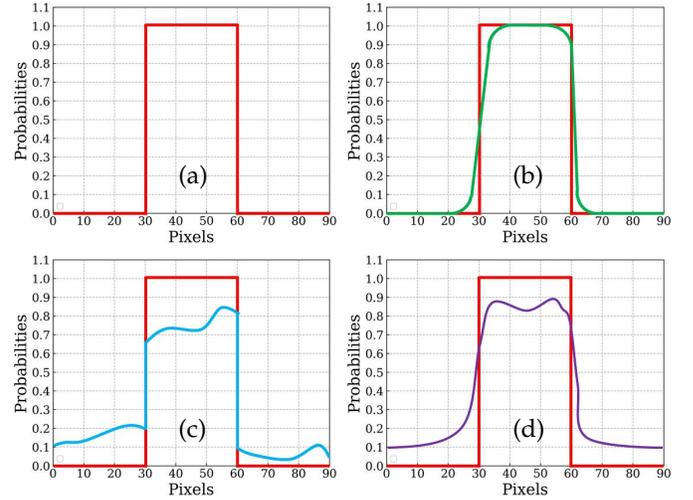
\begin{figure}[t!]
	\centering
	\begin{overpic}[width=\columnwidth]{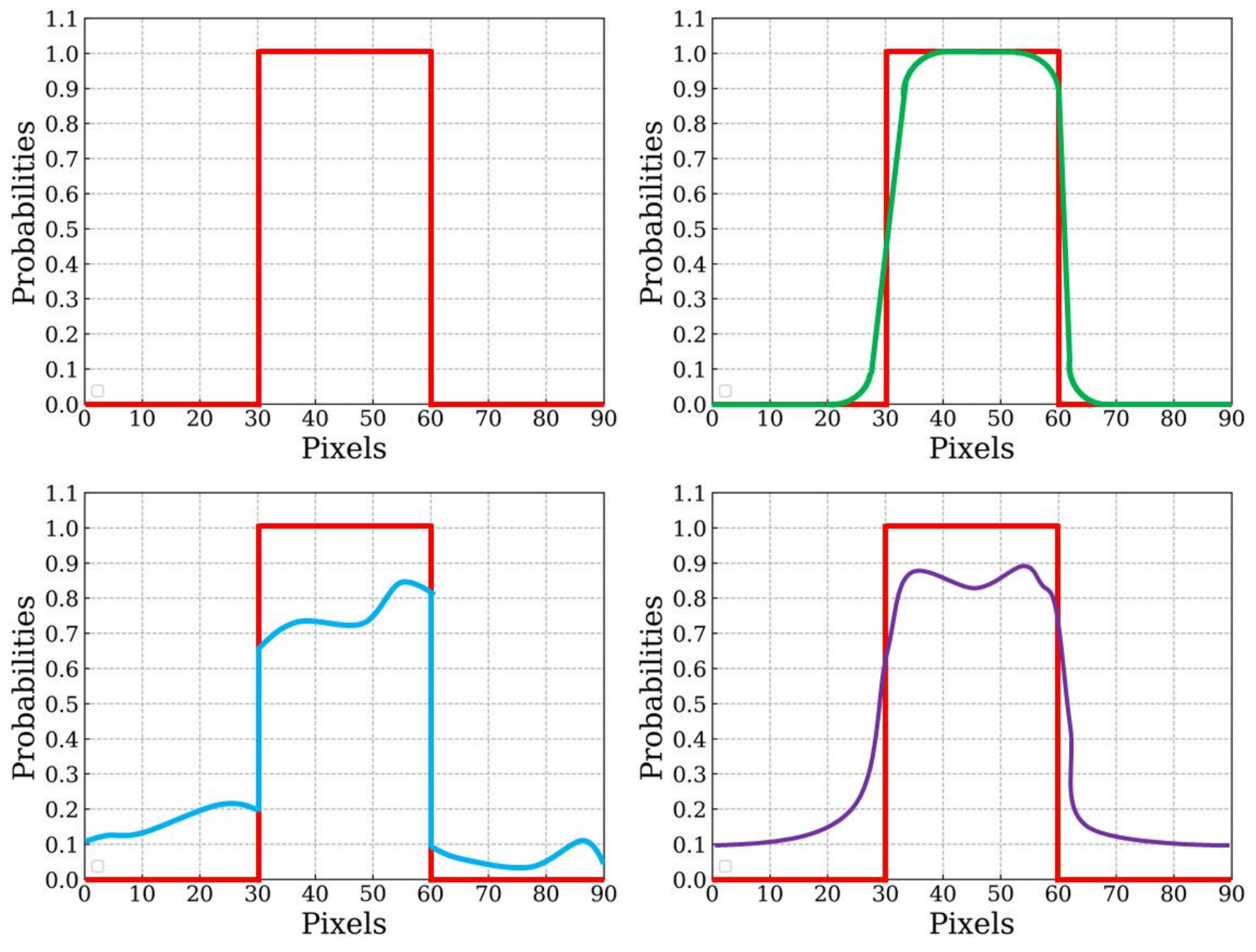}
	\put(25.5,48){(a)}
	\put(75.5,48){(b)}
	\put(25.5,10){(c)}
	\put(75.5,10){(d)}
  \end{overpic}
  \caption{Illustration of different aspects of coarse prediction in one-dimension: (a) Red: probability plot of GT, (b) Green: probability plot of coarse boundary not aligning with GT, (c) Blue: coarse region having too low probability, (d) Purple: real coarse predictions usually have both (b\&c) problems.
  }\label{fig:coarse}
\end{figure}

Our decoder is almost symmetrical to the encoder. 
Each stage consists of three convolution layers followed by a batch normalization and a ReLU activation function. 
The input of each stage is the concatenated feature maps of the up-sampled output from its previous stage and its corresponding stage in the encoder. 
To achieve the side-output maps, the multi-channel output of the bridge stage and each decoder stage is fed to a plain $3\times 3$ convolution layer followed by a bilinear upsampling and a sigmoid function. 
Therefore, given an input image, our prediction module produces seven segmentation probability maps in the training process. 
Although every predicted map is up-sampled to the same size with the input image, the last one has the highest accuracy and hence is taken as the final output of the prediction module. This output is passed to the refinement module.

\subsection{Residual Refinement Module}\label{sec:rm}
Refinement Modules (RMs) \cite{islamsalient, deng2018r3net} are usually designed as a residual block \cite{xu2017deep} that refines the coarse segmentation maps $S_{coarse}$ by learning the residuals $S_{residual}$ between the coarse maps and the GT, as 
\begin{equation}
    \centering
        S_{refined}=S_{coarse}+S_{residual}.
    \label{equ:rrm}
\end{equation} 
Before introducing our refinement module, the term ``coarse'' has to be determined. 
Here, ``coarse'' includes two aspects. 
One is blurry and noisy boundaries (see the one-dimension illustration in \figref{fig:coarse}(b)).
The other one is the unevenly predicted regional probabilities (see \figref{fig:coarse}(c)). 
As shown in \figref{fig:coarse}(d), real predicted coarse maps usually contain both coarse cases. 

\begin{figure}[t!]
\vspace{-5pt}
\begin{center}
    \subfigure[RRM\_LC]{\includegraphics[height=0.42\linewidth]{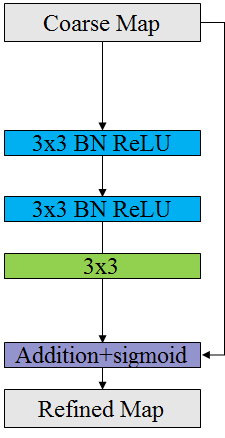}
	\label{fig:rf1}}
	\hfill
	\subfigure[RRM\_MS]{\includegraphics[height=0.42\linewidth]{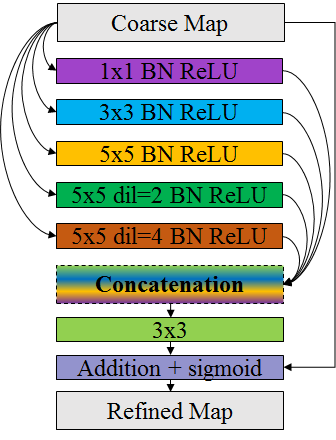}
	\label{fig:rf2}}
	\hfill
	\subfigure[RRM\_Ours]{\includegraphics[height=0.42\linewidth]{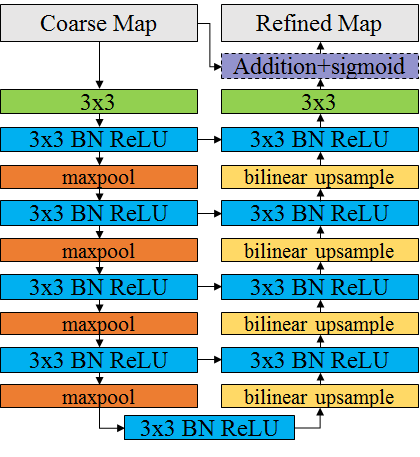}
	\label{fig:rf5}}
	\vspace{-5pt}
\end{center}
\caption{Illustration of different Residual Refine Modules (RRM): (a) local boundary refinement module RRM\_LC; (b) multi-scale refinement module RRM\_MS; (c) our encoder-decoder refinement module RRM\_Ours.}
\label{fig:rf}
\end{figure}

The residual refinement module based on local context (RRM\_LC), \figref{fig:rf1}, was originally designed for boundary refinement \cite{peng2017large}.
Since its receptive field is small, Islam \etal \cite{islamsalient} and Deng \etal \cite{deng2018r3net} iteratively or recurrently used it for refining segmentation probability maps on different scales. 
Wang \etal \cite{DBLP:conf/iccv/WangBZZL17} adopted the pyramid pooling module from \cite{he2014spatial}, in which three-scale pyramid pooling features are concatenated.
To avoid losing details caused by pooling operations, RRM\_MS 
( \figref{fig:rf2}) uses convolutions with different kernel sizes and dilations \cite{yu2015multi, zhang2018bi} to capture multi-scale contexts. 
However, these modules are shallow thus hard to capture high-level information for refinement.

To refine inaccuracies in coarse segmentation maps of image reigons and boundaries, 
we develop a novel residual refinement module. 
Our RRM employs the residual encoder-decoder architecture, RRM\_Ours (see Figs. \ref{fig:arc} and \ref{fig:rf5}). 
Its main architecture is similar but simpler than our prediction module. 
It contains an input layer, an encoder, a bridge, a decoder and an output layer. 
Different from the prediction module, both the encoder and decoder have four stages. 
Each stage only has one convolutional layer.
Each layer has 64 filters of size $3\times 3$ followed by a batch normalization and a ReLU activation function. 
The bridge stage also has a convolutional layer with 64 filters of size $3\times 3$ followed by a batch normalization and ReLU activation.
Non-overlapping max pooling is used for downsampling in the encoder and bilinear interpolation is utilized for upsampling in the decoder. 
The output of this RM module is used as the final generating segmentation results of BASNet.

\subsection{Hybrid Loss}
Our training loss is defined as the summation over all outputs:
\begin{equation}
    \centering
    \mathcal{L} = {\textstyle \sum}_{k=1}^K{\alpha_k \ell^{(k)}},
    \label{equ:loss}
\end{equation}
where $\ell^{(k)}$ is the loss of the $k$-th side output, $K$ denotes the total number of the outputs and  $\alpha_k$ is the weight of each loss. 
As described in Sec.~\ref{sec:pm} and Sec.~\ref{sec:rm}, our segmentation model is deeply supervised with eight outputs, i.e. $K=8$, including seven outputs from the prediction module and one output from the refinement module.

To obtain high quality regional segmentation and clear boundaries, we propose to define $\ell^{(k)}$ as a hybrid loss:
\begin{equation}
    \centering
    \ell^{(k)} = \ell_{bce}^{(k)} + \ell_{ssim}^{(k)} + \ell_{iou}^{(k)},
    \label{equ:lside}
\end{equation}
where $\ell_{bce}^{(k)}$,~$\ell_{ssim}^{(k)}$, and $\ell_{iou}^{(k)}$ denote BCE loss \cite{DBLP:journals/anor/BoerKMR05}, SSIM loss \cite{wang2003multiscale} and IoU loss \cite{mattyus2017deeproadmapper}, respectively.

BCE \cite{DBLP:journals/anor/BoerKMR05} loss is the most widely used loss in binary classification and segmentation. It is defined as:
\begin{equation}
    \centering
    {\scriptstyle
    \ell_{bce} = -\sum\limits_{(r,c)}[G(r,c)\log(S(r,c))+(1-G(r,c))\log(1-S(r,c))]},
    \label{equ:bce_loss}
\end{equation}
where $G(r,c)\in\{0,1\}$ is the GT label of the pixel $(r,c)$ and $S(r,c)$ is the predicted probability of segmented object. 

SSIM~\cite{wang2003multiscale} was originally devised for image quality assessment. 
It captures the structural information in an image. 
Hence, we integrated it into our training loss to learn the structural information of the GT. 
Let $\mathbf{x}=\{x_j:j=1,...,N^2\}$ and $\mathbf{y}=\{y_j:j=1,...,N^2\}$ be the pixel values of two corresponding patches (size: $N\times N$) cropped from the predicted probability map $S$ and the binary GT mask $G$, respectively. The SSIM of $\mathbf{x}$ and $\mathbf{y}$ is defined as:
\begin{equation}
    \centering
    \small{\ell_{ssim}=1 -  \frac{(2\mu_x\mu_y+C_1)(2\sigma_{xy}+C_2)}{(\mu_x^2+\mu_y^2+C_1)(\sigma_x^2+\sigma_y^2+C_2)}}
    \label{equ:ssim_loss}
\end{equation}
where $\mu_x$, $\mu_y$ and $\sigma_x$, $\sigma_y$ are the mean and standard deviations of $\mathbf{x}$ and $\mathbf{y}$ respectively, $\sigma_{xy}$ is covariance, $C_1=0.01^2$ and $C_2=0.03^2$ are used to avoid dividing by 0. 

IoU was originally proposed for measuring the similarity between two sets \cite{jaccard1912distribution} and has become a standard evaluation measure for object detection and segmentation. 
Recently, it has been used as a training loss \cite{rahman2016optimizing,mattyus2017deeproadmapper}. 
To ensure its differentiability, we adopted the IoU loss used in \cite{mattyus2017deeproadmapper}:
\begin{equation}
    \centering
    \ell_{iou} = 1 - \tfrac{\sum\limits_{r=1}\limits^H\sum\limits_{c=1}\limits^WS(r,c)G(r,c)}{\sum\limits_{r=1}\limits^H\sum\limits_{c=1}\limits^W[S(r,c)+G(r,c)-S(r,c)G(r,c)]}
    \label{equ:liou}
\end{equation}
where $G(r,c)\in\{0,1\}$ is the GT label of the pixel $(r,c)$ and $S(r,c)$ is the predicted probability of segmented object.

\begin{figure}[t!]
	\centering
	\begin{overpic}[width=\columnwidth]{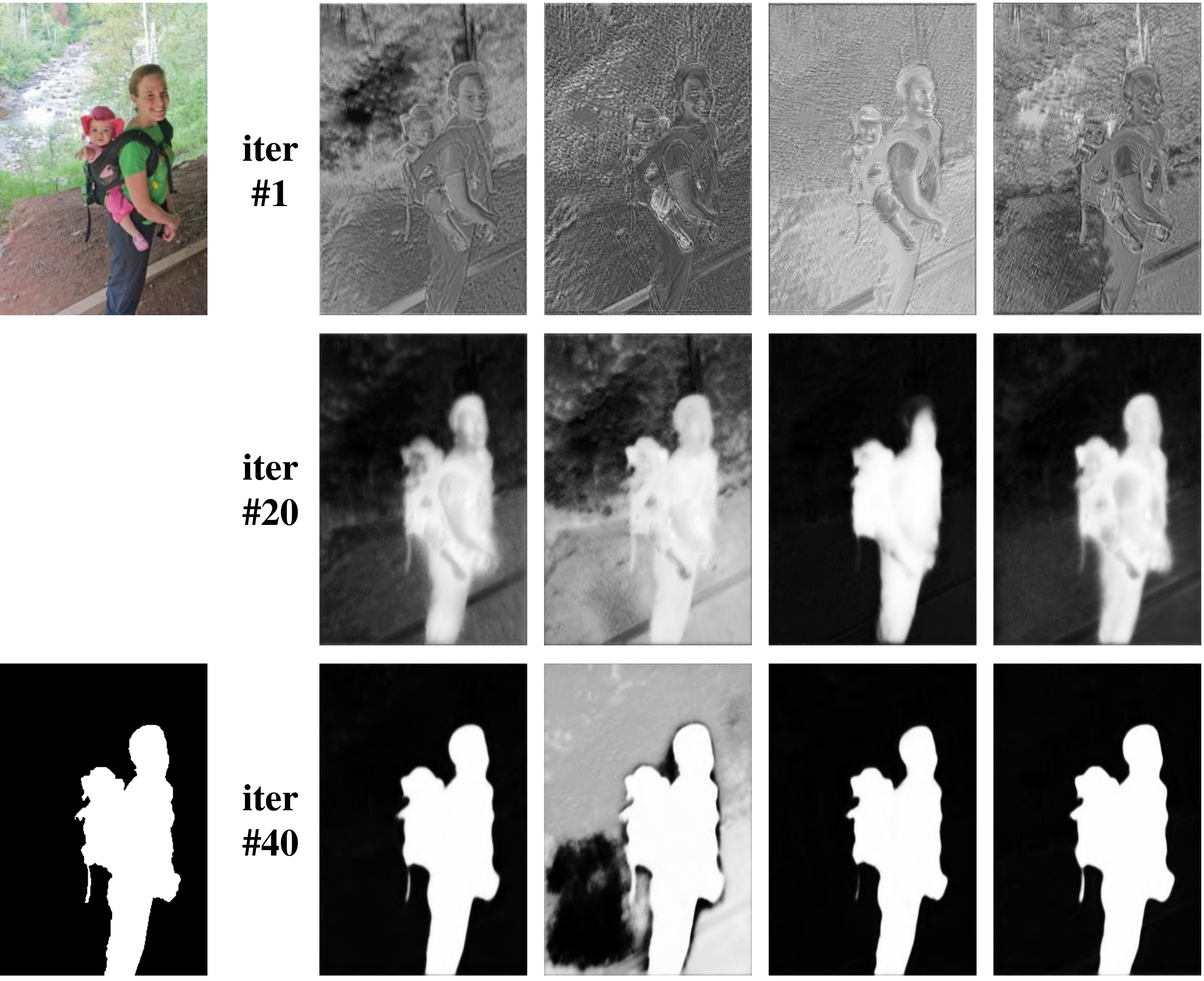}
	\put(1,51){\small (a) Image}
	\put(4,-4){\small (b) GT}
	\put(29.5,-4){\small (c) $\ell_{bce}$}
	\put(47,-4){\small (d) $\ell_{ssim}$}
	\put(67,-4){\small (e) $\ell_{iou}$}
	\put(86,-4){\small (f) $\ell_{bsi}$}
  \end{overpic}
  \vspace{5pt}
  \caption{Intermediate predictions of our BASNet when fitting with different losses.
  }\label{fig:hybrid_loss}
\end{figure}
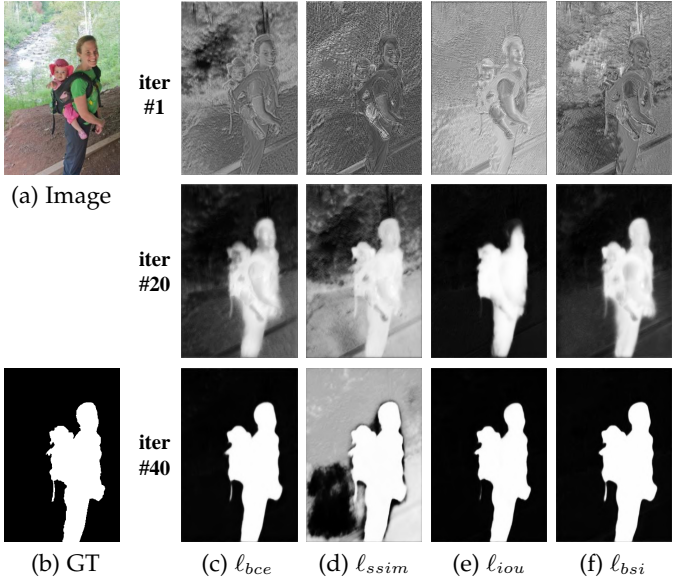

\figref{fig:hybrid_loss} illustrated the impact of each of the three losses. 
Fig.~\ref{fig:hybrid_loss} (a) and (b) are the input image and its ground truth segmentation mask. 
It is worth noting that the probability maps in Fig.~\ref{fig:hybrid_loss} are generated by fitting a single pair of image and its ground truth (GT) mask. Hence, after a certain number of iterations, all the losses are able to produce perfect results due to over-fitting. 
Here, we ignore the final fitting results and aim to observe the different characteristics and problems of these losses in the fitting process. 
The (c), (d), (e) and (f) columns show changes of the intermediate probability maps as the training progresses. 

The BCE loss is computed pixel-wise. It does not consider the labels of the neighborhood and it weights both the foreground and background pixels equally. 
This helps with the convergence on all pixels and guarantee a relatively good local optima. 
Since significantly erroneous predictions (predicting 0 as 0.9 or predicting 1 as 0.1) produce large BCE loss, the models trained with BCE loss suppress these errors by giving prediction values around 0.5 around the boundaries, which often leads to blurred boundaries and fine structures, as we can see from the second row of column (c), where the contour of the whole foreground region is blurring, and the third row, in which the cable below the backpack is with low probability values.   

The SSIM loss is a patch-level measure, which considers a local neighborhood of each pixel. 
It assigns higher weights to pixels located in the transitional buffer regions between foregrounds and backgrounds, \eg~boundaries, fine structures,  so that the loss is higher around the boundary, even when the predicted probabilities on the boundary and the rest of the foreground are the same. 
It is worth noting that the loss for the background region is similar or sometimes even higher than the foreground region. However, the background loss does not contribute to the training until the prediction of background pixel becomes very close to the GT, where the loss drops rapidly from one to zero. Because $\mu_y$, $\sigma_{xy}$, $\mu_x\mu_y$ and $\sigma_y^2$ in the SSIM loss (Equ. \ref{equ:ssim_loss}) are all zeros in the background regions, so the SSIM loss can be approximated by:
\begin{equation}
    \centering
    \small{\ell^{bg}_{ssim}=1 -  \frac{C_1C_2}{(\mu_x^2+C_1)(\sigma_x^2+C_2)}}.
    \label{equ:ssim_loss_aprx}
\end{equation}
Since $C_1=0.01^2$ and $C_2=0.03^2$, only if the prediction $x$ is close to zero, the SSIM loss (Equ. \ref{equ:ssim_loss_aprx}) will become the dominant term. 
The second and third rows of column (d) in Fig. \ref{fig:hybrid_loss} illustrate that the model trained with the SSIM loss is able to predict correct results on the foreground region and boundaries while neglecting the background accuracy in the beginning of the training process.  
This characteristic of the SSIM loss helps the optimization to focus on the boundary and foreground region. 
As the training progresses, the SSIM loss for the foreground is reduced and the background loss becomes the dominant term. 
This is helpful since the prediction typically goes close to zero only late in the training process, where BCE loss becomes flat. 
The SSIM loss ensures that there is still enough gradient to drive the learning process. 
Hence, the background prediction looks cleaner since the probability is pushed to zero. 

The IoU is a map-level measure. 
Larger ares contribute more to the IoU, so models trained with IoU loss emphasizes more on the large foreground regions and are thus able to produce relatively homogeneous and more confident (whiter) probabilities for these regions. However, these models often produce false negatives on fine structures. As shown in the column (e) of Fig.~\ref{fig:hybrid_loss}, the human head in the second row and the backpack cord in both the second and third rows are missing. 


To take advantage of the above three losses, we combine them together to formulate the hybrid loss. BCE is used to maintain a smooth gradient for all pixels, while IoU is employed to put more focus on the foreground. 
SSIM is used to encourage the prediction to respect the structure of the original image, by employing a larger loss near the boundaries, as well as further push the backgrounds predictions to zero.

\section{Experiments} 
In this paper, we are focusing on improving the spatial accuracy of segmentation results. Therefore, experiments are conducted on two reverse
binary class image segmentation tasks: salient object segmentation \cite{wang2017learning} and camouflaged object segmentation \cite{fan2020camouflaged}. 
Salient object segmentation is a popular task in computer vision, which aims at segmenting the salient regions against their backgrounds. In this task, the targets are usually with high contrast against their backgrounds. 
However, camouflaged object segmentation is the most challenging one because the camouflaged objects usually have similar appearance to their backgrounds, which means they are difficult to be perceived and segmented. In addition, many of the camouflaged objects have very complex structures and boundaries.



\subsection{Implementation and Setting}
We implement our network using the publicly available Pytorch 1.4.0 \cite{paszke2017automatic}. An 16-core PC with an AMD Threadripper 2950x 3.5 GHz CPU (with 64GB 3000 MHz RAM, ) and an RTX Titan GPU (with 24GB memory) is used for both training and testing. 
During training, each image is first resized to 320$\times$320 and randomly cropped to 288$\times$288. Some of the encoder parameters are initialized from the ResNet-34 model \cite{he2016deep}. Other convolutional layers are initialized by Xavier \cite{DBLP:journals/jmlr/GlorotB10}. We use the Adam optimizer \cite{kingma2014adam} to train our network and its hyperparameters are set to the default values, where the initial learning rate lr=1e-4, betas=(0.9, 0.999), eps=1e-8, weight\_decay=0. We train the network until the loss converges, without using the validation set. The training loss converges after 400k iterations with a batch size of eight and the whole training process takes about 110 hours. 
During testing, the input image is resized to 320$\times$320 and fed into the network to obtain its segmentation probability map. Then, the probability map (320$\times$320) is resized back to the original size of the input image. Both resizing processes use bilinear interpolation. The inference for a 320$\times$320 image only takes 0.015s (\textbf{70 fps}, different from that reported in our CVPR version~\cite{qin2019basnet}, in which IO time is included).


\subsection{Evaluation Metrics}\label{sec:metrics}
Five measures are used to evaluate the performance of the proposed model. 
\textbf{(1) Weighted F-measure} $F^w_\beta$ \cite{Margolin2014HowTE} gives a comprehensive and balanced evaluation on both precision and recall, which is able to better leverage the interpolation, dependency and equal-importance flaw. 
\textbf{(2) Relax boundary F-measure} $F^b_\beta$ \cite{ehrig2005relaxed} is adopted to quantitatively evaluate the boundary quality of the predicted maps. \textbf{(3) Mean absolute error} $M$ \cite{perazzi2012saliency} reflects the average per-pixel difference between the probability map and the GT. 
\textbf{(4) Mean structural measure} $S_\alpha$ \cite{DBLP:conf/iccv/FanCLLB17} quantizes the structural similarity between the predicted probability map and the GT mask. 
\textbf{(5) Mean enhanced-alignment measure} $E^m_\phi$ \cite{fan2018enhanced} takes both global and local similarity into consideration. Evaluation code: \url{https://github.com/DengPingFan/CODToolbox}.

\begin{figure*}[t!]
	\centering
	\begin{overpic}[width=\textwidth]{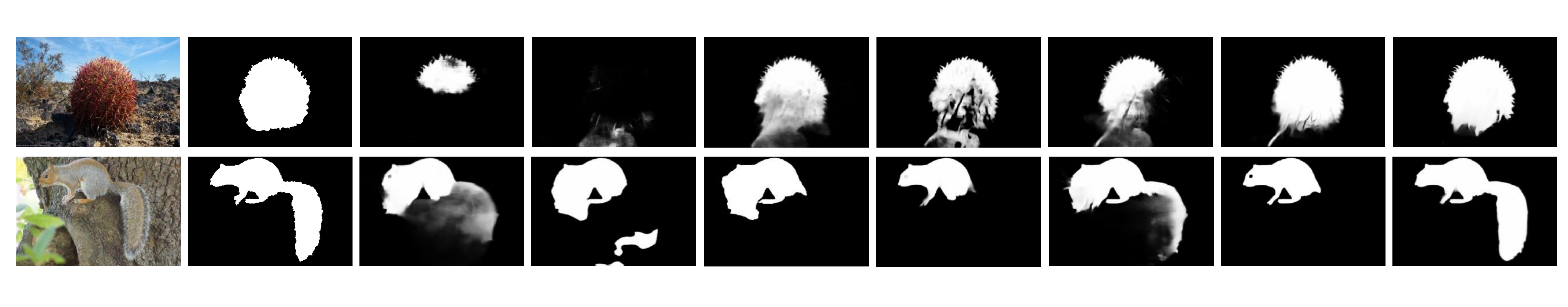}
	\put(3,18){\small (a) Image}
	\put(15,18){\small (b) GT}
	\put(23.5,18){\small (c) U-Net$_{bce}$}
	\put(36,18){\small (d) ED$_{bce}$}
	\put(46,18){\small (e) EDS$_{bce}$}
	\put(56,18){\small (f) ED\_LC$_{bce}$}
	\put(67,18){\small (g) ED\_MS$_{bce}$}
	\put(77.5,18){\small (h) BASNet$_{bce}$}
	\put(89,18){\small (i) BASNet$_{bsi}$}
	\put(3,0.5){\small (j) Image}
	\put(15,0.5){\small (k) GT}
	\put(26,0.5){\small (l) $\ell_{bce}$}
	\put(35.5,0.5){\small (m) $\ell_{ssim}$}
	\put(48,0.5){\small (n) $\ell_{iou}$}
	\put(59.5,0.5){\small (o) $\ell_{bs}$}
	\put(70,0.5){\small (p) $\ell_{bi}$}
	\put(81,0.5){\small (q) $\ell_{si}$}
	\put(92,0.5){\small (r) $\ell_{bsi}$}
  \end{overpic}
  \caption{Qualitative comparison of different configures in the ablation study. The first row show the predicted probability maps of different architectures trained with BCE loss and our BASNet trained with $\ell_{bsi}$ loss. The second row show the segmentation maps of our proposed prediction-refinement architecture trained with different losses. The corresponding quantitative results can be found in Table~\ref{tab:ablation}.
  }\label{fig:ab_arch_loss}
\end{figure*}

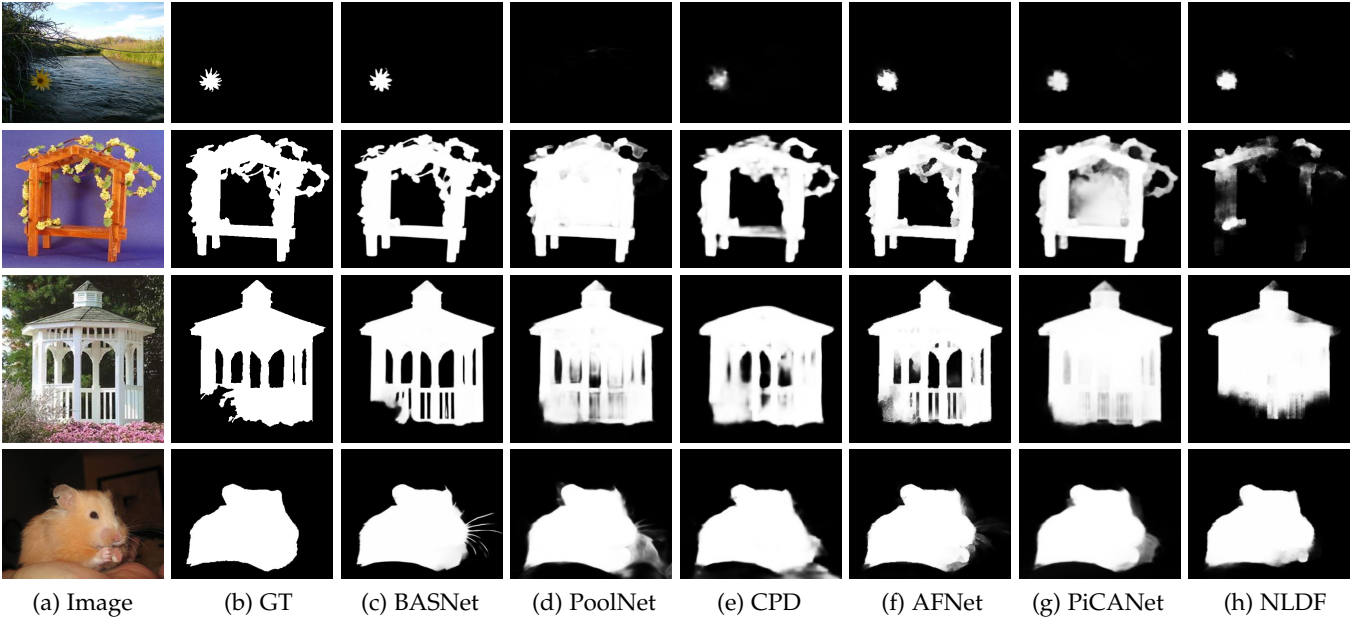
\begin{figure*}[t!]
	\centering
	\begin{overpic}[width=\textwidth]{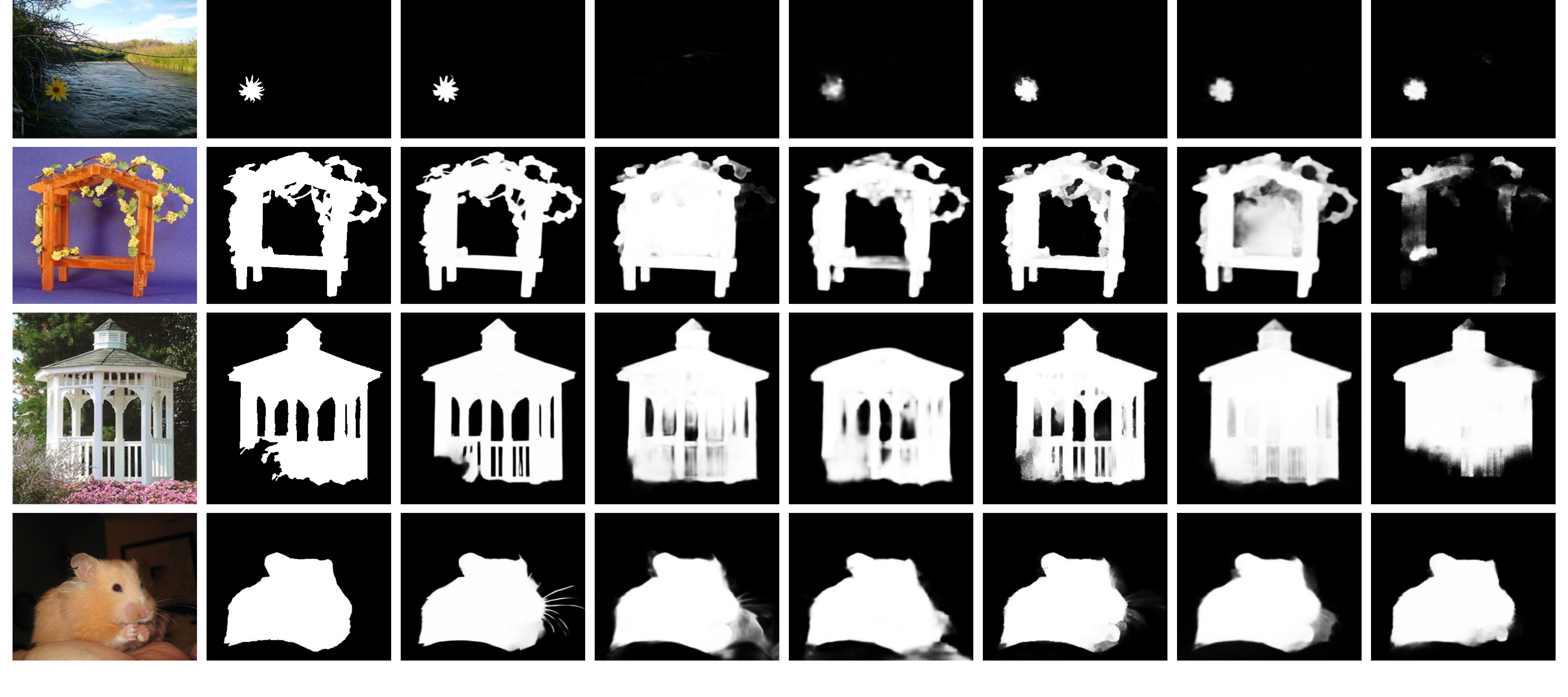}
	\put(3,0){\small (a) Image}
	\put(17,0){\small (b) GT}
	\put(27,0){\small (c) BASNet}
	\put(39.5,0){\small (d) PoolNet}
	\put(53,0){\small (e) CPD}
	\put(65,0){\small (f) AFNet}
	\put(76,0){\small (g) PiCANet}
	\put(90,0){\small (h) NLDF}
  \end{overpic}
  \caption{Qualitative comparison on salient object segmentation datasets.
  }\label{fig:qual_SOD}
\end{figure*}

\subsection{Experiments on Salient Object Segmentation}
\subsubsection{Datasets}
For salient object segmentation task\footnote{The camouflaged object segmentation task use the same augmentation strategies.}, we train our network using the DUTS-TR \cite{wang2017learning} dataset, which has 10553 images. Before training, the dataset is augmented by horizontal flipping to 21106 images. For salient object segmentation tasks, we evaluate our method on six commonly used salient object segmentation benchmark datasets: SOD \cite{movahedi2010design}, ECSSD \cite{yan2013hierarchical}, DUT-OMRON \cite{yang2013saliency}, PASCAL-S \cite{li2014secrets}, HKU-IS \cite{li2015visual}, DUTS-TE \cite{wang2017learning}. 
\textbf{DUT-OMRON} has 5,168 images with one or multiple objects. The majority of these objects are structurally complex. 
\textbf{PASCAL-S} was originally created for semantic image segmentation and consists of 850 challenging images. 
\textbf{DUTS} is a relatively large salient object segmentation dataset. It has two subsets: DUTS-TR and DUTS-TE. There are 10,553 images in \textbf{DUTS-TR} for training and 5,019 images in \textbf{DUTS-TE} for testing. In our experiments, DUTS-TR is used for training the model for salient object segmentation. 
\textbf{HKU-IS} contains 4,447 images, many of which contain multiple foreground objects. 
\textbf{ECSSD} contains 1,000 semantically meaningful images. However, the structures of the foreground objects in these images are complex. 
\textbf{SOD} contains 300 very challenging images. These images have either single complicated large foreground objects which overlap with the image boundaries or multiple salient objects with low contrast.

\subsubsection{Ablation Study}
In this section, we validate the effectiveness of each key components used in our model. 
The ablation study is divided into two parts: an architecture ablation and loss ablation. 
For simplify, the ablation experiments are conducted on the ECSSD dataset. The same hyper-parameters to that described in Sec. 4.1 are used here. 
\begin{table}[tbp] 
    \centering
    \footnotesize
    \setlength{\tabcolsep}{3.0pt}
    \renewcommand{\arraystretch}{0.5}
     \caption{Ablation study on different architectures (Arch.) and losses: ED: encoder-decoder, EDS: encoder-decoder + side output supervision; $\ell_{b}$, $\ell_{s}$ and $\ell_{i}$ denote the BCE, SSIM and IoU loss, respectively, $\ell_{bi}=\ell_{b}+\ell_{i}$, $\ell_{bs}=\ell_{b}+\ell_{s}$,
     $\ell_{si}=\ell_{s}+\ell_{i}$, $\ell_{bsi}=\ell_{b}+\ell_{s}+\ell_{i}$.}
    \label{tab:ablation}
    \begin{tabular}{c|c|ccccc}
		\hline
        Ablation & Configurations & $F^w_\beta\uparrow$ & $F^{b}_\beta\uparrow$  & $~M~\downarrow$ & $S_{\alpha}\uparrow$ & $E_{\phi}^{m}\uparrow$  \\
        \hline
        \multirow{6}{*}{\rotatebox[origin=c]{90}{\parbox[c]{2 cm}{\centering {\footnotesize Arch. }}}}
         & \multicolumn{1}{r|}{U-Net \cite{ronneberger2015u} + $\ell_{b}$} & 0.827 & 0.669 & 0.064 & 0.867 & 0.897\\
        & \multicolumn{1}{r|}{ED + $\ell_{b}$} & 0.871 & 0.786 & 0.045 & 0.908 & 0.923\\
        & \multicolumn{1}{r|}{EDS + $\ell_{b}$} & 0.891 & 0.819 & 0.041 & 0.920 & 0.935\\
        & \multicolumn{1}{r|}{EDS+RRM\_LC + $\ell_{b}$} & 0.900 & 0.804 & 0.038 & 0.915 & 0.935\\
        & \multicolumn{1}{r|}{EDS+RRM\_MS + $\ell_{b}$} & 0.890 & 0.816 & 0.041 & 0.919 & 0.934\\
        & \multicolumn{1}{r|}{EDS+RRM\_Ours + $\ell_{b}$} & 0.900 & 0.827 & 0.037 & 0.923 & 0.943\\
        \hline
        \multirow{6}{*}{\rotatebox[origin=c]{90}{\parbox[c]{2 cm}{\centering {\footnotesize Loss}}}}
 & \multicolumn{1}{l|}{EDS+RRM\_Ours + $\ell_{s}$} & 0.886 & 0.814 & 0.044 & 0.904 & 0.932\\
        & \multicolumn{1}{l|}{EDS+RRM\_Ours + $\ell_{i}$} & 0.902 & 0.820 & 0.037 & 0.911 & 0.943\\
        & \multicolumn{1}{l|}{EDS+RRM\_Ours + $\ell_{bs}$} & 0.903 & 0.823 & 0.037 & 0.920 & 0.942\\
        & \multicolumn{1}{l|}{EDS+RRM\_Ours + $\ell_{bi}$} & 0.909 & 0.832 & 0.035 & 0.921 & \bf{0.947}\\
        & \multicolumn{1}{l|}{EDS+RRM\_Ours + $\ell_{si}$} & 0.894 & 0.812 & 0.041 & 0.906 & 0.938\\
        & \multicolumn{1}{l|}{EDS+RRM\_Ours + $\ell_{bsi}$} & \bf{0.912} & \bf{0.840} & \bf{0.034} & \bf{0.925} & \bf{0.947} \\
        \hline
    \end{tabular}
\end{table}

\begin{table*}[t!]
	\caption{Comparison of the proposed method and 25 other methods on three salient object segmentation datasets: DUT-OMRON, DUTS-TE and HKU-IS. $\uparrow$ and $\downarrow$ indicate the higher the score the better and the lower the score the better, respectively. ``*'' indicates results post-processed by CRF. \textbf{Bold} font denotes the best performance.}
	\centering
  \footnotesize
  \renewcommand{\arraystretch}{0.5}
  \setlength\tabcolsep{5.5pt}
	\begin{tabular}{l|ccccc|ccccc|ccccc}
		\hline
		\multirow{2}{*}{\textbf{Models}} & \multicolumn{5}{|c|}{\textbf{DUT-OMRON}\cite{yang2013saliency}} & \multicolumn{5}{|c|}{\textbf{DUTS-TE}\cite{wang2017learning}} & \multicolumn{5}{|c}{\textbf{HKU-IS}\cite{li2015visual}}\\
		\cline{2-16}
		 & $F^w_\beta\uparrow$ & $F^{b}_\beta\uparrow$  & $~M~\downarrow$ & $S_{\alpha}\uparrow$ & $E_{\phi}^{m}\uparrow$ &
		  $F^w_\beta\uparrow$ & $F^{b}_\beta\uparrow$  & $M\downarrow$ & $S_{\alpha}\uparrow$ & $E_{\phi}^{m}\uparrow$ &
		  $F^w_\beta\uparrow$ & $F^{b}_\beta\uparrow$  & $M\downarrow$ & $S_{\alpha}\uparrow$ & $E_{\phi}^{m}\uparrow$\\
		 \hline 
        \textbf{MDF}$_\text{TIP16}$	&	0.565	&	0.406	&	0.142	&	0.721	&	0.759	&	0.543	&	0.447	&	0.099	&	0.723	&	0.764	&	0.564	&	0.594	&	0.129	&	0.81	&	0.742	\\
        \textbf{UCF}$_\text{ICCV17}$	&	0.573	&	0.48	&	0.12	&	0.76	&	0.761	&	0.596	&	0.518	&	0.112	&	0.777	&	0.776	&	0.779	&	0.679	&	0.062	&	0.875	&	0.887	\\
    \textbf{Amulet}$_\text{ICCV17}$	&	0.626	&	0.528	&	0.098	&	0.781	&	0.794	&	0.658	&	0.568	&	0.084	&	0.796	&	0.817	&	0.817	&	0.716	&	0.051	&	0.886	&	0.910	\\
    \textbf{NLDF$^*$}$_\text{CVPR17}$	&	0.634	&	0.514	&	0.08	&	0.77	&	0.799	&	0.71	&	0.591	&	0.065	&	0.805	&	0.851	&	0.838	&	0.694	&	0.048	&	0.879	&	0.914	\\
    \textbf{DSS$^*$}$_\text{CVPR17}$	&	0.697	&	0.559	&	0.063	&	0.79	&	0.831	&	0.755	&	0.606	&	0.056	&	0.812	&	0.877	&	0.867	&	0.706	&	0.04	&	0.878	&	0.925	\\
    \textbf{LFR}$_\text{IJCAI18}$	&	0.647	&	0.508	&	0.103	&	0.78	&	0.799	&	0.689	&	0.556	&	0.083	&	0.799	&	0.833	&	0.861	&	0.731	&	0.04	&	0.905	&	0.934	\\
    \textbf{C2S}$_\text{ECCV18}$	&	0.661	&	0.565	&	0.072	&	0.798	&	0.823	&	0.713	&	0.607	&	0.062	&	0.829	&	0.859	&	0.829	&	0.717	&	0.048	&	0.883	&	0.859	\\
    \textbf{RAS}$_\text{ECCV18}$	&	0.695	&	0.615	&	0.062	&	0.814	&	0.844	&	0.74	&	0.656	&	0.059	&	0.828	&	0.871	&	0.843	&	0.748	&	0.045	&	0.887	&	0.92	\\
    \textbf{RADF$^*$}$_\text{AAAI18}$	&	0.723	&	0.579	&	0.061	&	0.815	&	0.857	&	0.748	&	0.608	&	0.061	&	0.814	&	0.869	&	0.872	&	0.725	&	0.039	&	0.888	&	0.935	\\
    \textbf{PAGRN}$_\text{CVPR18}$	&	0.622	&	0.582	&	0.071	&	0.775	&	0.772	&	0.724	&	0.692	&	0.055	&	0.825	&	0.843	&	0.82	&	0.762	&	0.048	&	0.887	&	0.900	\\
    \textbf{BMPM}$_\text{CVPR18}$	&	0.681	&	0.612	&	0.064	&	0.809	&	0.831	&	0.761	&	0.699	&	0.048	&	0.851	&	0.883	&	0.859	&	0.773	&	0.039	&	0.907	&	0.931	\\
    \textbf{PiCANet}$_\text{CVPR18}$	&	0.691	&	0.643	&	0.068	&	0.826	&	0.833	&	0.747	&	0.704	&	0.054	&	0.851	&	0.873	&	0.847	&	0.784	&	0.042	&	0.906	&	0.923	\\
    \textbf{MLMS}$_\text{CVPR19}$	&	0.681	&	0.612	&	0.064	&	0.809	&	0.831	&	0.761	&	0.699	&	0.048	&	0.851	&	0.883	&	0.859	&	0.773	&	0.039	&	0.907	&	0.931	\\
    \textbf{AFNet}$_\text{CVPR19}$	&	0.717	&	0.635	&	0.057	&	0.826	&	0.846	&	0.785	&	0.714	&	0.046	&	0.855	&	0.893	&	0.869	&	0.772	&	0.036	&	0.905	&	0.935	\\
    \textbf{MSWS}$_\text{CVPR19}$	&	0.527	&	0.362	&	0.109	&	0.756	&	0.729	&	0.586	&	0.376	&	0.908	&	0.749	&	0.742	&	0.685	&	0.438	&	0.084	&	0.818	&	0.787	\\
    \textbf{R}$^3$\textbf{Net$^*$}$_\text{IJCAI18}$	&	0.728	&	0.599	&	0.063	&	0.817	&	0.853	&	0.763	&	0.601	&	0.058	&	0.817	&	0.873	&	0.877	&	0.74	&	0.036	&	0.895	&	0.939	\\
    \textbf{CapSal}$_\text{CVPR19}$	&	0.482	&	0.396	&	0.101	&	0.674	&	0.659	&	0.691	&	0.605	&	0.072	&	0.808	&	0.849	&	0.782	&	0.654	&	0.062	&	0.85	&	0.883	\\
    \textbf{SRM}$_\text{ICCV17}$	&	0.658	&	0.523	&	0.069	&	0.798	&	0.808	&	0.722	&	0.592	&	0.058	&	0.824	&	0.853	&	0.835	&	0.68	&	0.046	&	0.887	&	0.913	\\
    \textbf{DGRL}$_\text{CVPR18}$	&	0.697	&	0.584	&	0.063	&	0.810	&	0.845	&	0.76	&	0.656	&	0.051	&	0.836	&	0.887	&	0.865	&	0.744	&	0.037	&	0.897	&	0.939	\\
    \textbf{CPD}$_\text{CVPR19}$	&	0.719	&	0.655	&	\textbf{0.056}	&	0.825	&	0.847	&	0.795	&	0.741	&	0.043	&	0.858	&	0.898	&	0.875	&	0.795	&	0.034	&	0.905	&	0.939	\\
    \textbf{PoolNet}$_\text{CVPR19}$	&	0.729	&	0.675	&	\textbf{0.056}	&	0.836	&	0.854	&	0.807	&	0.765	&	\textbf{0.040}	&	0.871	&	0.904	&	0.881	&	0.811	&	0.033	&	0.917	&	0.940	\\
    \textbf{BANet}$_\text{ICCV19}$	&	0.719	&	0.611	&	0.061	&	0.823	&	0.861	&	0.781	&	0.687	&	0.046	&	0.861	&	0.897	&	0.869	&	0.760	&	0.037	&	0.902	&	0.938	\\
    \textbf{EGNet}$_\text{ICCV19}$	&	0.728	&	0.679	&	\textbf{0.056}	&	0.836	&	0.853	&	0.797	&	0.761	&	0.043	&	0.879	&	0.898	&	0.875	&	0.802	&	0.035	&	0.910	&	0.938	\\
    \textbf{MINet}$_\text{CVPR20}$	&	0.719	&	0.640	&	0.057	&	0.822	&	0.846	&	0.813	&	0.747	&	\textbf{0.040}	&	0.875	&	0.906	&	0.889	&	0.799	&	0.032	&	0.912	&	0.944	\\
    \textbf{GateNet}$_\text{ECCV20}$	&	0.703	&	0.625	&	0.061	&	0.821	&	0.840	&	0.786	&	0.722	&	0.045	&	0.871	&	0.892	&	0.872	&	0.783	&	0.036	&	0.910	&	0.934	\\
    \hline
    \textbf{BASNet} (Ours)	&	\textbf{0.760}	&	\textbf{0.703}	&	0.057	&	\textbf{0.841}	&	\textbf{0.868}	&	\textbf{0.825}	&	\textbf{0.786}	&	0.042	&	\textbf{0.881}	&	\textbf{0.907}	&	\textbf{0.900}	& \textbf{0.821}	&	\textbf{0.030}	&	\textbf{0.918}	&	\textbf{0.948}	\\
	\hline
	\end{tabular}
	\label{tab:compSOA}
\end{table*}

\begin{table*}[t!]
	\caption{Comparison of the proposed method and 25 other methods on three salient object detection datasets: ECSSD, PASCAL-S and SOD. See \tabref{tab:compSOA} for details.
	}
	\centering
	\footnotesize
    \renewcommand{\arraystretch}{0.5}
    \setlength\tabcolsep{5.2pt}
	\begin{tabular}{l|ccccc|ccccc|ccccc}
		\hline
		\multirow{2}{*}{\textbf{Baseline Models}} & \multicolumn{5}{|c|}{\textbf{ECSSD}\cite{yan2013hierarchical}} & \multicolumn{5}{|c|}{\textbf{PASCAL-S}\cite{li2014secrets}} & \multicolumn{5}{|c}{\textbf{SOD}\cite{movahedi2010design}}\\
		\cline{2-16}
		 & $F^w_\beta\uparrow$ & $F^{b}_\beta\uparrow$  & $~M~\downarrow$ & $S_{\alpha}\uparrow$ & $E_{\phi}^{m}\uparrow$ &
		  $F^w_\beta\uparrow$ & $F^{b}_\beta\uparrow$  & $M\downarrow$ & $S_{\alpha}\uparrow$ & $E_{\phi}^{m}\uparrow$ &
		  $F^w_\beta\uparrow$ & $F^{b}_\beta\uparrow$  & $M\downarrow$ & $S_{\alpha}\uparrow$ & $E_{\phi}^{m}\uparrow$\\
		 \hline 
		 \textbf{MDF}$_\text{TIP16}$	&	0.705	&	0.472	&	0.105	&	0.776	&	0.796	&	0.589	&	0.343	&	0.142	&	0.696	&	0.706	&	0.508	&	0.311	&	0.192	&	0.643	&	0.607	\\
        \textbf{UCF}$_\text{ICCV17}$	&	0.806	&	0.669	&	0.069	&	0.884	&	0.891	&	0.694	&	0.493	&	0.115	&	0.805	&	0.809	&	0.675	&	0.471	&	0.148	&	0.762	&	0.773	\\
        \textbf{Amulet}$_\text{ICCV17}$	&	0.84	&	0.711	&	0.059	&	0.894	&	0.909	&	0.734	&	0.541	&	0.100	&	0.818	&	0.835	&	0.677	&	0.454	&	0.144	&	0.753	&	0.776	\\
        \textbf{NLDF$^*$}$_\text{CVPR17}$	&	0.839	&	0.666	&	0.063	&	0.897	&	0.900	&	0.737	&	0.495	&	0.098	&	0.798	&	0.839	&	0.709	&	0.475	&	0.125	&	0.755	&	0.777	\\
        \textbf{DSS$^*$}$_\text{CVPR17}$	&	0.872	&	0.696	&	0.052	&	0.882	&	0.918	&	0.759	&	0.499	&	0.093	&	0.798	&	0.845	&	0.710	&	0.444	&	0.124	&	0.743	&	0.774	\\
        \textbf{LFR}$_\text{IJCAI18}$	&	0.858	&	0.694	&	0.052	&	0.897	&	0.923	&	0.737	&	0.499	&	0.107	&	0.805	&	0.835	&	0.734	&	0.479	&	0.123	&	0.773	&	0.813	\\
        \textbf{C2S}$_\text{ECCV18}$	&	0.851	&	0.708	&	0.055	&	0.893	&	0.917	&	0.766	&	0.543	&	0.082	&	0.836	&	0.864	&	0.700	&	0.457	&	0.124	&	0.760	&	0.785	\\
        \textbf{RAS}$_\text{ECCV18}$	&	0.857	&	0.741	&	0.056	&	0.893	&	0.914	&	0.736	&	0.560	&	0.101	&	0.799	&	0.830	&	0.720	&	0.544	&	0.124	&	0.764	&	0.788	\\
        \textbf{RADF$^*$}$_\text{AAAI18}$	&	0.883	&	0.720	&	0.049	&	0.894	&	0.929	&	0.755	&	0.515	&	0.097	&	0.802	&	0.840	&	0.729	&	0.476	&	0.126	&	0.757	&	0.801	\\
        \textbf{PAGRN}$_\text{CVPR18}$	&	0.834	&	0.747	&	0.061	&	0.889	&	0.895	&	0.738	&	0.594	&	0.090	&	0.822	&	0.830	&	-	&	-	&	-	&	-	&	-	\\
        \textbf{BMPM}$_\text{CVPR18}$	&	0.871	&	0.770	&	0.045	&	0.911	&	0.928	&	0.779	&	0.617	&	0.074	&	0.845	&	0.872	&	0.726	&	0.562	&	0.108	&	0.786	&	0.799	\\
        \textbf{PiCANet}$_\text{CVPR18}$	&	0.865	&	0.784	&	0.046	&	0.914	&	0.924	&	0.772	&	0.612	&	0.078	&	0.848	&	0.866	&	0.722	&	0.572	&	0.103	&	0.789	&	0.796	\\
        \textbf{MLMS}$_\text{CVPR19}$	&	0.871	&	0.770	&	0.045	&	0.911	&	0.928	&	0.779	&	0.62	&	0.074	&	0.844	&	0.875	&	0.726	&	0.562	&	0.108	&	0.786	&	0.799	\\
        \textbf{AFNet}$_\text{CVPR19}$	&	0.887	&	0.776	&	0.042	&	0.914	&	0.936	&	0.798	&	0.626	&	0.070	&	0.849	&	0.883	&	0.723	&	0.545	&	0.111	&	0.774	&	0.79	\\
        \textbf{MSWS}$_\text{CVPR19}$	&	0.716	&	0.411	&	0.096	&	0.828	&	0.791	&	0.614	&	0.289	&	0.133	&	0.768	&	0.731	&	0.573	&	0.231	&	0.167	&	0.700	&	0.656	\\
        \textbf{R}$^3$\textbf{Net$^*$}$_\text{IJCAI18}$	&	0.902	&	0.759	&	0.040	&	0.910	&	0.944	&	0.761	&	0.538	&	0.092	&	0.807	&	0.843	&	0.735	&	0.431	&	0.125	&	0.759	&	0.796	\\
        \textbf{CapSal}$_\text{CVPR19}$	&	0.771	&	0.574	&	0.077	&	0.826	&	0.849	&	0.786	&	0.527	&	0.073	&	0.837	&	0.872	&	0.597	&	0.404	&	0.148	&	0.695	&	0.699	\\
        \textbf{SRM}$_\text{ICCV17}$	&	0.853	&	0.672	&	0.054	&	0.895	&	0.913	&	0.758	&	0.509	&	0.084	&	0.834	&	0.853	&	0.670	&	0.392	&	0.128	&	0.741	&	0.744	\\
        \textbf{DGRL}$_\text{CVPR18}$	&	0.883	&	0.753	&	0.042	&	0.906	&	0.938	&	0.787	&	0.569	&	0.074	&	0.839	&	0.877	&	0.731	&	0.502	&	0.106	&	0.773	&	0.807	\\
        \textbf{CPD}$_\text{CVPR19}$	&	0.898	&	0.811	&	0.037	&	0.918	&	0.942	&	0.800	&	0.639	&	0.071	&	0.848	&	0.878	&	0.714	&	0.556	&	0.112	&	0.767	&	0.778	\\
        \textbf{PoolNet}$_\text{CVPR19}$	&	0.896	&	0.813	&	0.039	&	0.921	&	0.940	&	0.798	&	0.644	&	0.075	&	0.832	&	0.876	&	0.759	&	0.606	&	0.102	&	\textbf{0.797}	&	0.818	\\
        \textbf{BANet}$_\text{ICCV19}$	&	0.890	&	0.758	&	0.041	&	0.913	&	0.940	&	0.792	&	0.589	&	0.078	&	0.840	&	0.875	&	0.750	&	0.589	&	0.109	&	0.782	&	0.813	\\
        \textbf{EGNet}$_\text{ICCV19}$	&	0.892	&	0.814	&	0.041	&	0.920	&	0.936	&	0.793	&	0.650	&	0.077	&	0.848	&	0.873	&	0.737	&	0.586	&	0.112	&	0.784	&	0.798	\\
        \textbf{MINet}$_\text{CVPR20}$	&	0.905	&	0.805	&	0.037	&	0.920	&	0.943	&	\textbf{0.813}	&	0.648	&	\textbf{0.065}	&	0.854	&	\textbf{0.889}	&	-	&	-	&	-	&	-	&	-	\\
        \textbf{GateNet}$_\text{ECCV20}$	&	0.886	&	0.782	&	0.042	&	0.917	&	0.933	&	0.803	&	0.623	&	0.068	&	\textbf{0.857}	&	0.882	&	-	&	-	&	-	&	-	&	-	\\
        \hline
        \textbf{BASNet} (Ours)	&	\textbf{0.912}	&	\textbf{0.840}	&	\textbf{0.034}	&	\textbf{0.925}	&	\textbf{0.947}	&	0.808	&	\textbf{0.674}	&	0.072	&	0.847	&	0.878	&	\textbf{0.762}	&	\textbf{0.640}	&	\textbf{0.102}	& 0.793	&	\textbf{0.822}	\\	
		\hline
	\end{tabular}
	\label{tab:compSOB}
\end{table*}

\textbf{Architecture:} To demonstrate the effectiveness of our BASNet, we report quantitative comparison results of our model against other related architectures.  
We take U-Net \cite{ronneberger2015u} as our baseline network. 
Then we start with our proposed encoder-decoder network and progressively extend it with dense side output supervision and different residual refinement modules, including RRM\_LC, RRM\_MS and RRM\_Ours. 
The top part of Table~\ref{tab:ablation} and the first row of Fig.~\ref{fig:ab_arch_loss} illustrate the qualitative and quantitative results of the architecture ablation study, respectively. 
As we can see, our BASNet architecture achieves the best performance among all configurations.

\begin{figure*}[t!]
	\centering
	\begin{overpic}[width=\textwidth]{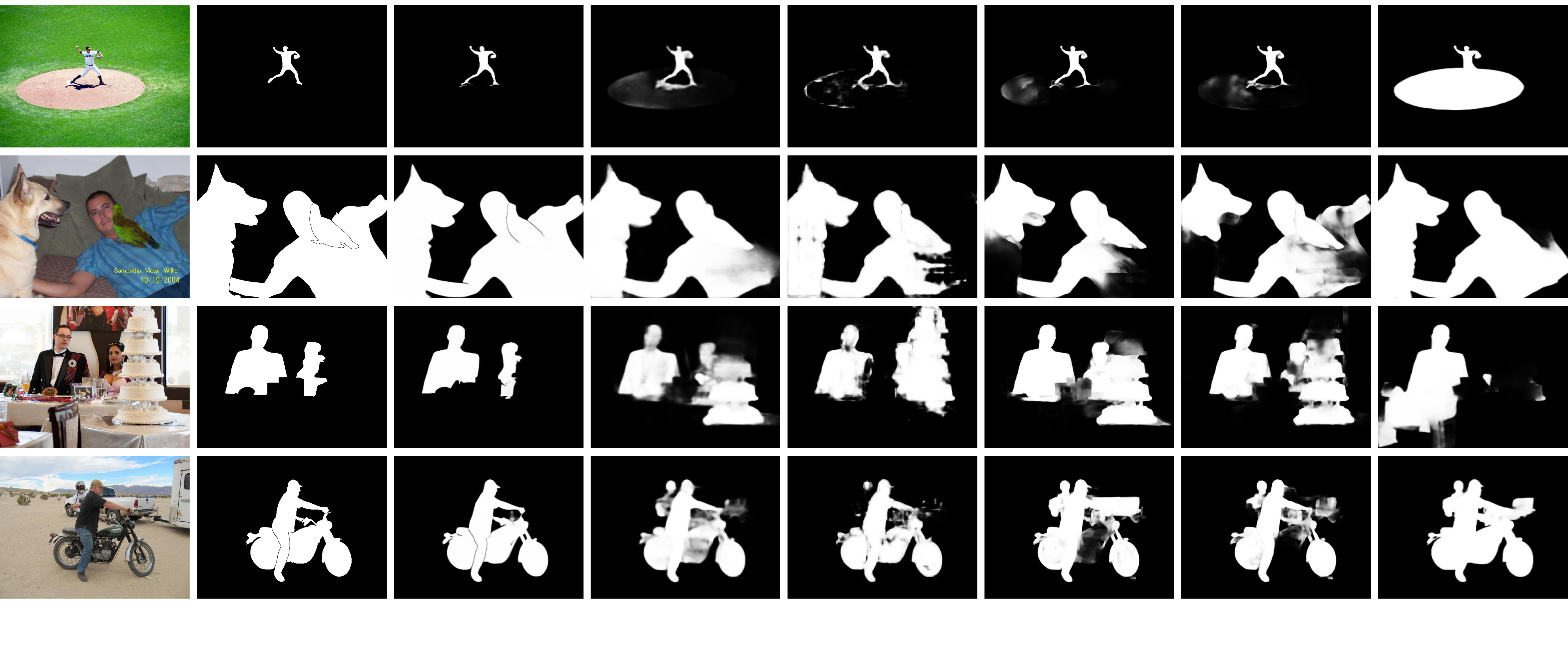}
	\put(3,1){\small (a) Image}
	\put(17,1){\small (b) GT}
	\put(27,1){\small (c) BASNet}
	\put(39.5,1){\small (d) SCRN}
	\put(53,1){\small (e) BANet}
	\put(65,1){\small (f) EGNet}
	\put(76,1){\small (g) PoolNet}
	\put(90,1){\small (h) CPD}
  \end{overpic}
  \caption{Qualitative comparison on typical samples from the SOC dataset. Images from top to bottom are from attributes SO (Small Object), OV(Out-of-View), OC (Occlusion) and SC (Shape Complexity) respectively.
  }\label{fig:qual-soc}
\end{figure*}

\begin{figure}[t!]
	\centering
	\begin{overpic}[width=.9\columnwidth]{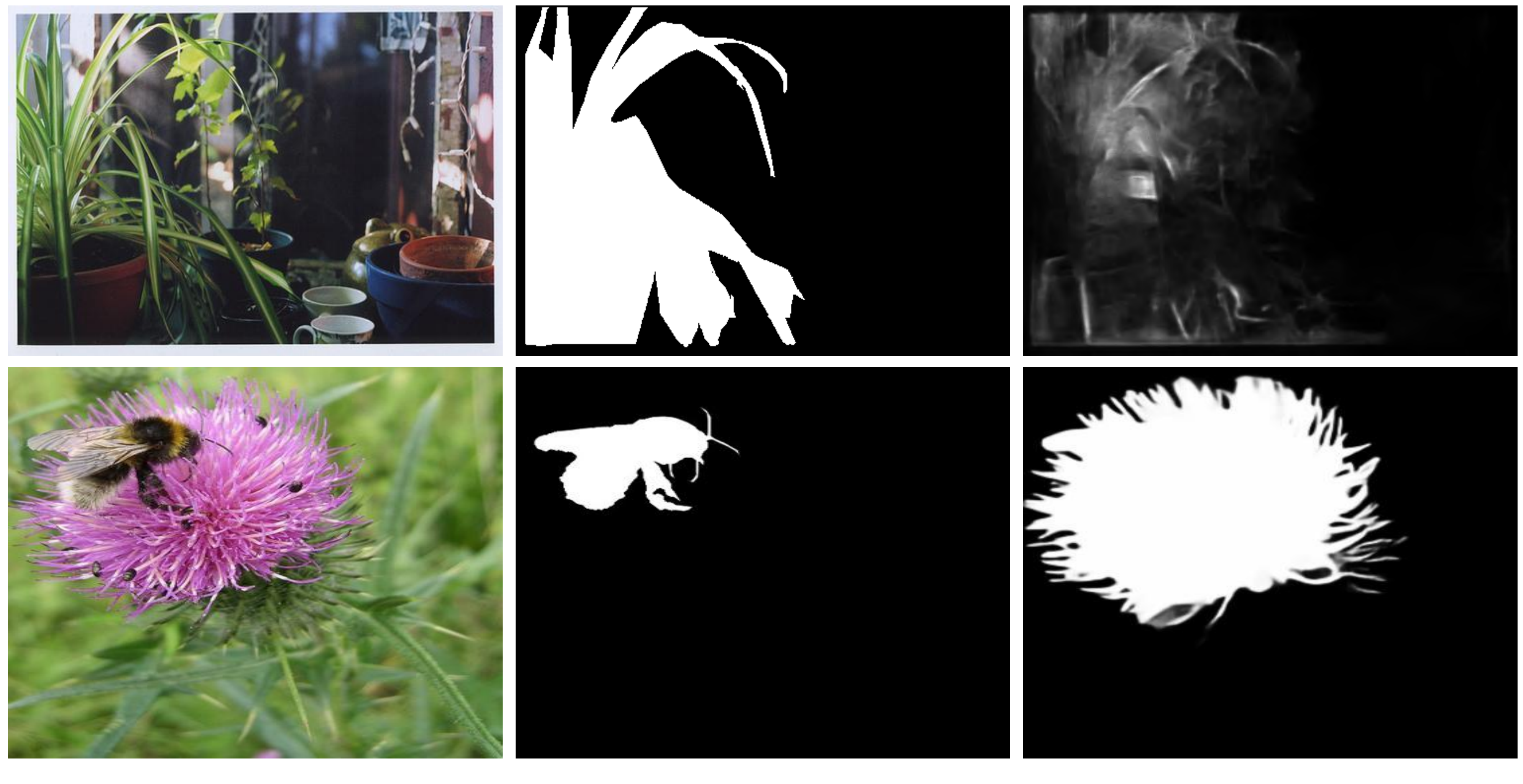}
	\put(9,-3){\small (a) Image}
	\put(45,-3){\small (b) GT}
	\put(75,-3){\small (c) BASNet}
  \end{overpic}
  \vspace{2pt}
  \caption{Failure cases on salient object segmentation datasets.
  }\label{fig:fail_sod}
\end{figure}

\textbf{Loss:} To demonstrate the effectiveness of our proposed fusion loss, we conduct a set of experiments over different losses based on our BASNet architecture. 
The results in Table \ref{tab:ablation} indicate that the proposed hybrid $\ell_{bsi}$ loss greatly improves the performance, especially in terms of the boundary quality. It is clear that our hybrid loss achieves superior qualitative results, as shown in the second row of Fig.~\ref{fig:ab_arch_loss}.

\subsubsection{Comparison with State-of-the-Arts}
We compare our method  with 25 state-of-the-art models, including MDF \cite{li2016visual}, 
UCF \cite{DBLP:conf/iccv/ZhangWLWY17},
Amulet \cite{DBLP:conf/iccv/ZhangWLWR17}, 
NLDF \cite{luo2017non}, 
DSS \cite{hou2017deeply}, 
LFR \cite{DBLP:conf/ijcai/ZhangLLS18}, 
C2S \cite{DBLP:conf/eccv/LiYCLS18}, 
RAS \cite{DBLP:conf/eccv/ChenTWH18}, 
RADF \cite{DBLP:conf/aaai/HuZQFH18}, 
PAGRN \cite{zhang2018progressive}, 
BMPM \cite{zhang2018bi}, 
PiCANet \cite{liu2018picanet}, 
MLMS \cite{MLMS}, 
AFNet \cite{AFNet}, 
MSWS \cite{MSWS}, 
R$^3$-Net \cite{deng2018r3net}, 
CapSal \cite{CapSal}, 
SRM \cite{DBLP:conf/iccv/WangBZZL17}, 
DGRL \cite{wang2018detect}, 
CPD \cite{wu2019cascaded}, 
PoolNet \cite{liu2019simple}, 
BANet \cite{su2019selectivity}, 
EGNet \cite{zhao2019egnet}, 
MINet \cite{pang2020multi} and 
GateNet \cite{GateNet}, on the salient object segmentation task. 
For fair comparison, we either use the segmentation maps released by the authors or run their publicly available models with their default settings.

\textbf{Quantitative Evaluation:}
Tables \ref{tab:compSOA} and \ref{tab:compSOB} provide quantitative comparisons on six salient object segmentation datasets. Our BASNet outperforms other models on the DUT-OMRON, DUTS-TE, HKU-IS, ECSSD and SOD datasets in terms of nearly all metrics, except for the $M$ measures on DUT-OMRON and DUTS-TE and the $S_\alpha$ on SOD. On the PASCAL-S dataset, MINet performs the best in terms of three metrics: $F^w_\beta$, $MAE$ and $E^m_\phi$. It is worth noting that BASNet achieves the highest relax boundary F-measure $F^b_\beta$ on all of the six datasets, which indicates its strong capability in capturing boundaries and fine structures. 

\textbf{Qualitative Evaluation:}
Fig.~\ref{fig:qual_SOD} shows the qualitative comparison between our BASNet and 5 other typical models. As we can see, our BASNet is able to handle different challenging cases, such as small target with relatively low contrast (1st row), large object with complicated boundaries (2nd row), object with hollow structures (3rd row) and target with very fine structures (4th row). The third and fourth row of Fig.~\ref{fig:qual_SOD} show inspiring results, in which the segmentation maps predicted by our BASNet contain more details than the GT. These details reveals the possible inconsistency between the labels of training and testing datasets. Although detecting of these details usually leads to the deterioration of the quantitative evaluation scores of our model, it is more practically useful than good scores. 

\subsubsection{Failure Cases}
Fig. \ref{fig:fail_sod} shows three typical failure cases of our BASNet on SOD datasets. For instance, the model sometimes fails in very complicated scenarios, in which there seems no salient objects, as show in the first row of Fig. \ref{fig:fail_sod}. 
The second row gives an exemplary failure case of ``saliency confusing'', where the scene contains multiple independent ``salient'' targets, But only one of them is labeled. Our BASNet sometimes fails in these cases due to lack of the ability of recognizing the tiny saliency differences between multiple connected targets. The recent uncertainty model~\cite{zhang2020uncertainty} may be one of the solutions.

\begin{table*}[tbp]
	\caption{Comparison of the proposed method and other SOTA methods on the SOC test set. $\uparrow$ and $\downarrow$ indicate the higher score the better and the lower score the better respectively. \textbf{Bold} font indicates the best performance. \textbf{Avg.} denotes the average of all the attribute-based metric scores.}
	\centering
    \centering
	\footnotesize
    \renewcommand{\arraystretch}{0.5}
    \setlength\tabcolsep{0.2pt}
	\begin{tabular}{l|c|cccccccccccccccc|cc}
		\hline
		\textbf{Attr}	&	\textbf{Metr.}	&	\tabincell{c}{\textbf{Amulet}\\\cite{DBLP:conf/iccv/ZhangWLWR17}}	&	\tabincell{c}{\textbf{DSS}\\\cite{hou2017deeply}}	&	\tabincell{c}{\textbf{NLDF}\\\cite{luo2017non}}	&	\tabincell{c}{\textbf{C2SNet}\\\cite{DBLP:conf/eccv/LiYCLS18}}	&	\tabincell{c}{\textbf{SRM}\\\cite{DBLP:conf/iccv/WangBZZL17}}	&	\tabincell{c}{\textbf{R3Net}\\\cite{deng2018r3net}}	&	\tabincell{c}{\textbf{BMPM}\\\cite{zhang2018bi}}	&	\tabincell{c}{\textbf{DGRL}\\\cite{wang2018detect}}	&	\tabincell{c}{\textbf{PiC(R)}\\\cite{liu2018picanet}}	&	\tabincell{c}{\textbf{RANet}\\\cite{chen2020reverse}}	&	\tabincell{c}{\textbf{AFNet}\\\cite{AFNet}}	&	\tabincell{c}{\textbf{CPD}\\\cite{wu2019cascaded}}	&	\tabincell{c}{\textbf{PoolNet}\\\cite{liu2019simple}}	&	\tabincell{c}{\textbf{EGNet}\\\cite{zhao2019egnet}}	&	\tabincell{c}{\textbf{BANet}\\\cite{su2019selectivity}}	&	\tabincell{c}{\textbf{SCRN}\\\cite{Wu_2019_ICCV}}	& 
		\tabincell{c}{\textbf{Ours}\\(DUTS)} & 
		\tabincell{c}{\textbf{Ours}\\(SOC)}	\\
		\hline 
		\multirow{5}{*}{\textbf{AC}}	&	$F^w_\beta\uparrow$	&	0.620	&	0.629	&	0.620	&	0.647	&	0.690	&	0.593	&	0.680	&	0.718	&	0.682	&	0.603	&	0.712	&	0.727	&	0.713	&	0.731	&	0.740	&	0.724	&	0.735	&	\textbf{0.792}	\\
	    &	$F^{b}_\beta\uparrow$	&	0.448	&	0.384	&	0.374	&	0.408	&	0.410	&	0.387	&	0.531	&	0.457	&	0.489	&	0.448	&	0.569	&	0.626	&	0.578	&	0.597	&	0.562	&	0.588	&	0.659	&	\textbf{0.696}	\\
	    &	$~M~\downarrow$	&	0.120	&	0.113	&	0.119	&	0.109	&	0.096	&	0.135	&	0.098	&	0.081	&	0.093	&	0.132	&	0.084	&	0.083	&	0.094	&	0.085	&	0.086	&	0.078	&	0.087	&	\textbf{0.060}	\\
	    &	$S_{\alpha}\uparrow$	&	0.752	&	0.753	&	0.737	&	0.755	&	0.791	&	0.713	&	0.780	&	0.790	&	0.792	&	0.708	&	0.796	&	0.799	&	0.795	&	0.806	&	0.806	&	0.809	&	0.805	&	\textbf{0.831}	\\
	    &	$E_{\phi}^{m}\uparrow$	&	0.791	&	0.788	&	0.784	&	0.807	&	0.824	&	0.753	&	0.815	&	0.853	&	0.815	&	0.765	&	0.852	&	0.843	&	0.846	&	0.854	&	0.858	&	0.849	&	0.844	&	\textbf{0.885}	\\
	\hline
    \multirow{5}{*}{\textbf{BO}}	&	$F^w_\beta\uparrow$	&	0.612	&	0.614	&	0.622	&	0.730	&	0.667	&	0.456	&	0.670	&	0.786	&	0.799	&	0.453	&	0.741	&	0.739	&	0.610	&	0.585	&	0.720	&	0.778	&	0.747	&	\textbf{0.808}	\\
	&	$F^{b}_\beta\uparrow$	&	0.274	&	0.213	&	0.218	&	0.362	&	0.274	&	0.229	&	0.400	&	0.392	&	0.466	&	0.231	&	0.450	&	0.481	&	0.323	&	0.319	&	0.360	&	0.453	&	0.519	&	\textbf{0.572}	\\
	&	$~M~\downarrow$	&	0.346	&	0.356	&	0.354	&	0.267	&	0.306	&	0.445	&	0.303	&	0.215	&	0.200	&	0.454	&	0.245	&	0.257	&	0.353	&	0.373	&	0.271	&	0.224	&	0.253	&	\textbf{0.166}	\\
	&	$S_{\alpha}\uparrow$	&	0.574	&	0.561	&	0.568	&	0.654	&	0.614	&	0.437	&	0.604	&	0.684	&	0.729	&	0.421	&	0.658	&	0.647	&	0.561	&	0.528	&	0.645	&	0.698	&	0.666	&	\textbf{0.723}	\\
	&	$E_{\phi}^{m}\uparrow$	&	0.551	&	0.537	&	0.539	&	0.661	&	0.616	&	0.419	&	0.620	&	0.725	&	0.741	&	0.404	&	0.698	&	0.665	&	0.554	&	0.528	&	0.650	&	0.706	&	0.677	&	\textbf{0.775}	\\
	\hline														
    \multirow{5}{*}{\textbf{CL}}	&	$F^w_\beta\uparrow$	&	0.663	&	0.617	&	0.614	&	0.655	&	0.665	&	0.546	&	0.678	&	0.714	&	0.692	&	0.542	&	0.696	&	0.724	&	0.681	&	0.677	&	0.726	&	0.717	&	0.700	&	\textbf{0.730}	\\
	&	$F^{b}_\beta\uparrow$	&	0.374	&	0.275	&	0.292	&	0.342	&	0.327	&	0.315	&	0.432	&	0.393	&	0.420	&	0.344	&	0.465	&	0.553	&	0.488	&	0.493	&	0.461	&	0.506	&	0.552	&	\textbf{0.579}	\\
	&	$~M~\downarrow$	&	0.141	&	0.153	&	0.159	&	0.144	&	0.134	&	0.182	&	0.123	&	0.119	&	0.123	&	0.188	&	0.119	&	0.114	&	0.134	&	0.139	&	0.117	&	0.113	&	0.121	&	\textbf{0.110}	\\
	&	$S_{\alpha}\uparrow$	&	0.763	&	0.722	&	0.713	&	0.742	&	0.759	&	0.659	&	0.761	&	0.770	&	0.787	&	0.624	&	0.768	&	0.773	&	0.760	&	0.757	&	0.784	&	\textbf{0.795}	&	0.774	&	0.785	\\
	&	$E_{\phi}^{m}\uparrow$	&	0.789	&	0.763	&	0.764	&	0.789	&	0.793	&	0.710	&	0.801	&	0.824	&	0.794	&	0.715	&	0.802	&	0.821	&	0.801	&	0.790	&	0.824	&	0.820	&	0.807	&	\textbf{0.826}	\\
	\hline														
    \multirow{5}{*}{\textbf{HO}}	&	$F^w_\beta\uparrow$	&	0.688	&	0.660	&	0.661	&	0.668	&	0.696	&	0.633	&	0.684	&	0.722	&	0.704	&	0.626	&	0.722	&	0.751	&	0.739	&	0.720	&	0.754	&	0.743	&	0.746	&	\textbf{0.764}	\\
	&	$F^{b}_\beta\uparrow$	&	0.465	&	0.347	&	0.378	&	0.398	&	0.392	&	0.383	&	0.496	&	0.447	&	0.462	&	0.425	&	0.527	&	0.618	&	0.570	&	0.560	&	0.542	&	0.577	&	0.627	&	\textbf{0.639}	\\
	&	$~M~\downarrow$	&	0.119	&	0.124	&	0.126	&	0.123	&	0.115	&	0.136	&	0.116	&	0.104	&	0.108	&	0.143	&	0.103	&	0.097	&	0.100	&	0.106	&	0.094	&	0.096	&	0.099	&	\textbf{0.093}	\\
	&	$S_{\alpha}\uparrow$	&	0.791	&	0.767	&	0.755	&	0.768	&	0.794	&	0.740	&	0.781	&	0.791	&	0.809	&	0.713	&	0.798	&	0.803	&	0.815	&	0.802	&	0.819	&	\textbf{0.823}	&	0.809	&	0.814	\\
	&	$E_{\phi}^{m}\uparrow$	&	0.810	&	0.796	&	0.798	&	0.805	&	0.819	&	0.782	&	0.813	&	0.833	&	0.819	&	0.777	&	0.834	&	0.845	&	0.846	&	0.829	&	\textbf{0.850}	&	0.842	&	0.843	&	\textbf{0.850}	\\
	\hline															\multirow{5}{*}{\textbf{MB}}	&	$F^w_\beta\uparrow$	&	0.561	&	0.577	&	0.551	&	0.593	&	0.619	&	0.489	&	0.651	&	0.655	&	0.637	&	0.576	&	0.626	&	0.679	&	0.642	&	0.649	&	0.672	&	0.690	&	0.678	&	\textbf{0.725}	\\
	&	$F^{b}_\beta\uparrow$	&	0.435	&	0.396	&	0.397	&	0.450	&	0.395	&	0.348	&	0.561	&	0.464	&	0.520	&	0.476	&	0.537	&	0.619	&	0.592	&	0.584	&	0.539	&	0.595	&	0.635	&	\textbf{0.674}	\\
	&	$~M~\downarrow$	&	0.142	&	0.132	&	0.138	&	0.128	&	0.115	&	0.160	&	0.105	&	0.113	&	0.099	&	0.139	&	0.111	&	0.106	&	0.121	&	0.109	&	0.104	&	0.100	&	0.115	&	\textbf{0.072}	\\
	&	$S_{\alpha}\uparrow$	&	0.712	&	0.719	&	0.685	&	0.720	&	0.742	&	0.657	&	0.762	&	0.744	&	0.775	&	0.696	&	0.734	&	0.754	&	0.751	&	0.762	&	0.764	&	0.792	&	0.755	&	\textbf{0.797}	\\
	&	$E_{\phi}^{m}\uparrow$	&	0.739	&	0.753	&	0.740	&	0.778	&	0.778	&	0.697	&	0.812	&	0.823	&	0.813	&	0.761	&	0.762	&	0.804	&	0.779	&	0.789	&	0.803	&	0.817	&	0.805	&	\textbf{0.836}	\\
	\hline															\multirow{5}{*}{\textbf{OC}}	&	$F^w_\beta\uparrow$	&	0.607	&	0.595	&	0.593	&	0.622	&	0.630	&	0.520	&	0.644	&	0.659	&	0.638	&	0.527	&	0.680	&	0.672	&	0.659	&	0.658	&	0.678	&	0.673	&	0.672	&	\textbf{0.707}	\\
	&	$F^{b}_\beta\uparrow$	&	0.395	&	0.310	&	0.335	&	0.382	&	0.343	&	0.323	&	0.456	&	0.396	&	0.439	&	0.382	&	0.503	&	0.545	&	0.510	&	0.505	&	0.466	&	0.514	&	0.573	&	\textbf{0.601}	\\
	&	$~M~\downarrow$	&	0.143	&	0.144	&	0.149	&	0.130	&	0.129	&	0.168	&	0.119	&	0.116	&	0.119	&	0.169	&	0.109	&	0.115	&	0.119	&	0.121	&	0.112	&	0.111	&	0.115	&	\textbf{0.101}	\\
	&	$S_{\alpha}\uparrow$	&	0.735	&	0.718	&	0.709	&	0.738	&	0.749	&	0.653	&	0.752	&	0.747	&	0.765	&	0.641	&	0.771	&	0.750	&	0.756	&	0.754	&	0.765	&	0.775	&	0.760	&	\textbf{0.780}	\\
	&	$E_{\phi}^{m}\uparrow$	&	0.763	&	0.760	&	0.755	&	0.784	&	0.780	&	0.706	&	0.800	&	0.808	&	0.784	&	0.718	&	0.820	&	0.810	&	0.801	&	0.798	&	0.809	&	0.800	&	0.806	&	\textbf{0.829}	\\
	\hline															\multirow{5}{*}{\textbf{OV}}	&	$F^w_\beta\uparrow$	&	0.637	&	0.622	&	0.616	&	0.671	&	0.682	&	0.527	&	0.701	&	0.733	&	0.721	&	0.529	&	0.723	&	0.721	&	0.697	&	0.707	&	\textbf{0.752}	&	0.723	&	0.730	&	0.749	\\
	&	$F^{b}_\beta\uparrow$	&	0.405	&	0.311	&	0.339	&	0.420	&	0.368	&	0.336	&	0.494	&	0.434	&	0.490	&	0.383	&	0.524	&	0.592	&	0.526	&	0.541	&	0.509	&	0.545	&	0.617	&	\textbf{0.630}	\\
	&	$~M~\downarrow$	&	0.173	&	0.180	&	0.184	&	0.159	&	0.150	&	0.216	&	0.136	&	0.125	&	0.127	&	0.217	&	0.129	&	0.134	&	0.148	&	0.146	&	0.119	&	0.126	&	0.132	&	\textbf{0.114}	\\
	&	$S_{\alpha}\uparrow$	&	0.721	&	0.700	&	0.688	&	0.728	&	0.745	&	0.624	&	0.751	&	0.762	&	\textbf{0.781}	&	0.611	&	0.761	&	0.748	&	0.747	&	0.752	&	0.779	&	0.774	&	0.764	&	\textbf{0.781}	\\
	&	$E_{\phi}^{m}\uparrow$	&	0.751	&	0.737	&	0.736	&	0.790	&	0.779	&	0.663	&	0.807	&	0.828	&	0.810	&	0.664	&	0.817	&	0.803	&	0.795	&	0.802	&	\textbf{0.835}	&	0.808	&	0.809	&	0.828	\\
	\hline															\multirow{5}{*}{\textbf{SC}}	&	$F^w_\beta\uparrow$	&	0.608	&	0.599	&	0.593	&	0.611	&	0.638	&	0.550	&	0.677	&	0.669	&	0.627	&	0.594	&	0.696	&	0.708	&	0.695	&	0.678	&	0.706	&	0.691	&	0.728	&	\textbf{0.746}	\\
	&	$F^{b}_\beta\uparrow$	&	0.481	&	0.407	&	0.414	&	0.433	&	0.423	&	0.427	&	0.561	&	0.455	&	0.492	&	0.504	&	0.572	&	0.627	&	0.613	&	0.597	&	0.562	&	0.603	&	0.654	&	\textbf{0.672}	\\
	&	$~M~\downarrow$	&	0.098	&	0.098	&	0.101	&	0.100	&	0.090	&	0.114	&	0.081	&	0.087	&	0.093	&	0.110	&	0.076	&	0.080	&	0.075	&	0.083	&	0.078	&	0.078	&	0.074	&	\textbf{0.072}	\\
	&	$S_{\alpha}\uparrow$	&	0.768	&	0.761	&	0.745	&	0.756	&	0.783	&	0.716	&	0.799	&	0.772	&	0.784	&	0.724	&	0.808	&	0.793	&	0.807	&	0.793	&	0.807	&	0.809	&	0.812	&	\textbf{0.820}	\\
	&	$E_{\phi}^{m}\uparrow$	&	0.794	&	0.799	&	0.788	&	0.806	&	0.814	&	0.765	&	0.841	&	0.837	&	0.799	&	0.792	&	0.854	&	0.858	&	0.856	&	0.844	&	0.851	&	0.843	&	0.861	&	\textbf{0.872}	\\
	\hline															\multirow{5}{*}{\textbf{SO}} &	$F^w_\beta\uparrow$	&	0.523	&	0.524	&	0.526	&	0.531	&	0.561	&	0.487	&	0.567	&	0.602	&	0.566	&	0.518	&	0.596	&	0.623	&	0.626	&	0.594	&	0.621	&	0.614	&	0.634	&	\textbf{0.684}	\\
	&	$F^{b}_\beta\uparrow$	&	0.386	&	0.325	&	0.341	&	0.353	&	0.334	&	0.342	&	0.442	&	0.382	&	0.417	&	0.412	&	0.468	&	0.533	&	0.523	&	0.494	&	0.457	&	0.506	&	0.551	&	\textbf{0.612}	\\
	&	$~M~\downarrow$	&	0.119	&	0.109	&	0.115	&	0.116	&	0.099	&	0.118	&	0.096	&	0.092	&	0.095	&	0.113	&	0.089	&	0.091	&	0.087	&	0.098	&	0.090	&	0.082	&	0.092	&	\textbf{0.075}	\\
	&	$S_{\alpha}\uparrow$	&	0.718	&	0.713	&	0.703	&	0.706	&	0.737	&	0.682	&	0.732	&	0.736	&	0.748	&	0.682	&	0.746	&	0.745	&	0.768	&	0.749	&	0.755	&	0.767	&	0.758	&	\textbf{0.787}	\\
	&	$E_{\phi}^{m}\uparrow$	&	0.745	&	0.756	&	0.747	&	0.752	&	0.769	&	0.732	&	0.780	&	0.802	&	0.766	&	0.759	&	0.792	&	0.804	&	0.814	&	0.784	&	0.801	&	0.797	&	0.800	&	\textbf{0.835}	\\
	\hline															\multirow{5}{*}{\textbf{Avg.}}	&	$F^w_\beta\uparrow$	&	0.613	&	0.604	&	0.600	&	0.636	&	0.650	&	0.533	&	0.661	&	0.695	&	0.674	&	0.552	&	0.688	&	0.705	&	0.674	&	0.667	&	0.708	&	0.706	&	0.708	&	\textbf{0.745}	\\
	&	$F^{b}_\beta\uparrow$	&	0.407	&	0.330	&	0.343	&	0.394	&	0.363	&	0.343	&	0.486	&	0.424	&	0.466	&	0.401	&	0.513	&	0.577	&	0.525	&	0.521	&	0.495	&	0.543	&	0.599	&	\textbf{0.631}	\\
	&	$~M~\downarrow$	&	0.156	&	0.157	&	0.161	&	0.142	&	0.137	&	0.186	&	0.131	&	0.117	&	0.117	&	0.185	&	0.118	&	0.120	&	0.137	&	0.140	&	0.119	&	0.112	&	0.121	&	\textbf{0.096}	\\
	&	$S_{\alpha}\uparrow$	&	0.726	&	0.713	&	0.700	&	0.730	&	0.746	&	0.653	&	0.747	&	0.755	&	0.774	&	0.647	&	0.760	&	0.757	&	0.751	&	0.745	&	0.769	&	0.782	&	0.767	&	\textbf{0.791}	\\
	&	$E_{\phi}^{m}\uparrow$	&	0.748	&	0.743	&	0.739	&	0.775	&	0.775	&	0.692	&	0.788	&	0.815	&	0.793	&	0.706	&	0.803	&	0.806	&	0.788	&	0.780	&	0.809	&	0.809	&	0.806	&	\textbf{0.837}	\\
		\hline
	\end{tabular}
	\label{tab:compSOC}
\end{table*}

\subsubsection{Attribute-base Analysis}
In addition to the most frequently used salient object segmentation datasets, we also test our model on another dataset, SOC \cite{fan2018SOC}. The SOC dataset contains complicated scenarios, which are more challenging than those in the previous six SOD datasets. Besides, the SOC dataset categorizes images into nine different groups including AC (Appearance Change), BO (Big Object), CL (Clutter), HO (Heterogeneous Object), MB (Motion Blur), OC (Occlusion), OV (Out-of-View), SC (Shape Complexity), and SO (Small Object),  according to their attributes. We train our BASNet on both DUTS-TR and the training set (1,800 images with salient objects) of SOC dataset \cite{fan2018SOC} and evaluate their performance on the testing set of SOC-Sal. There are totally 600 images with salient objects in the testing set. Each image may be categorized into one or multiple attributes (\eg AC and BO).

\textbf{Quantitative Evaluation:} 
Tab. \ref{tab:compSOC} illustrates a comparison between our BASNet and 16 other state-of-the-art models, including Amulet \cite{DBLP:conf/iccv/ZhangWLWR17}, DSS \cite{hou2017deeply}, NLDF \cite{luo2017non}, C2S-Net \cite{DBLP:conf/eccv/LiYCLS18}, SRM \cite{DBLP:conf/iccv/WangBZZL17}, R3Net \cite{deng2018r3net}, BMPM \cite{zhang2018bi}, DGRL \cite{wang2018detect}, PiCANet-R (PiC(R)) \cite{liu2018picanet}, RANet \cite{chen2020reverse}, AFNet \cite{AFNet}, CPD \cite{wu2019cascaded}, PoolNet \cite{liu2019simple}, EGNet \cite{zhao2019egnet}, BANet \cite{su2019selectivity} and SCRN \cite{Wu_2019_ICCV} in terms of attribute-based performance. As we can see, our BASNet achieves obvious improvements against the existing methods. Particularly, our BASNet advances the boundary measure $F^b_\beta$ by large margins (over $5\%$ and sometimes over $10\%$) on different attributes. 

\textbf{Qualitative Evaluation:} Fig.~\ref{fig:qual-soc} provides a qualitative comparison of our BASNet and other baseline models. As we can see, BASNet is able to handle different challenges, including small objects (1st row), out-of-view objects (2nd row), occluded targets (3rd row) and objects with complicated shapes (4th row).

\subsection{Experiments on Camouflaged Object Segmentation}
To further evaluate the performance of the proposed BASNet, we also tested it on the camouflaged object segmentation (COS) task \cite{Jinnan-2020,ltnghia-AAAI2021,fan2020camouflaged}. Compared with salient object segmentation, COS is a relatively newer and more challenging task. Because the contrast between the camouflaged targets and their backgrounds is sometimes extremely low. Besides, the targets usually have similar color and texture to their backgrounds. In addition, their shape or structure of these targets can sometimes be very complex. 

\begin{table*}[t!]
	\caption{Comparison of the proposed method and 13 other methods on three camouflaged object segmentation datasets: CHAMELEON, CAMO-Test and COD10K-Test. $\uparrow$ and $\downarrow$ indicate the higher score the better and the lower the score the better, respectively. \textbf{Bold} font indicates the best performance.}
	\centering
    \centering
	\footnotesize
    \renewcommand{\arraystretch}{0.5}
    \setlength\tabcolsep{5.0pt}
	\begin{tabular}{l|ccccc|ccccc|ccccc}
		\hline
		\multirow{2}{*}{\textbf{Baseline Models}} & \multicolumn{5}{|c|}{\textbf{CHAMELEON}\cite{chameleon}} & \multicolumn{5}{|c|}{\textbf{CAMO-Test}\cite{le2019anabranch}} & \multicolumn{5}{|c}{\textbf{COD10K-Test}\cite{fan2020camouflaged}}\\
		\cline{2-16}
		 & $F^w_\beta\uparrow$ & $F^{b}_\beta\uparrow$  & $~M~\downarrow$ & $S_{\alpha}\uparrow$ & $E_{\phi}^{m}\uparrow$ &
		 $F^w_\beta\uparrow$ & $F^{b}_\beta\uparrow$  & $M\downarrow$ & $S_{\alpha}\uparrow$ & $E_{\phi}^{m}\uparrow$ &
		 $F^w_\beta\uparrow$ & $F^{b}_\beta\uparrow$  & $M\downarrow$ & $S_{\alpha}\uparrow$ & $E_{\phi}^{m}\uparrow$\\
		 \hline 
		 \textbf{FPN}$_\text{CVPR17}$	 &	0.590    &	0.246	&	0.075	&	0.794	&	0.784	&	0.483	&	0.232	&	0.131	&	0.684	&	0.677	&	0.411	&	0.195	&	0.075	&	0.697	&	0.692	\\
        \textbf{MaskRCNN}$_\text{CVPR17}$	&	0.518	&	0.128	&	0.099	&	0.643	&	0.778	&	0.430	&	0.117	&	0.151	&	0.574	&	0.715	&	0.402	&	0.110	&	0.081	&	0.613	&	0.748	\\
        \textbf{PSPNet}$_\text{CVPR17}$	&	0.555	&	0.207	&	0.085	&	0.773	&	0.756	&	0.455	&	0.191	&	0.139	&	0.663	&	0.659	&	0.377	&	0.166	&	0.080	&	0.678	&	0.681	\\
        \textbf{UNet++}$_\text{DLMIA18}$	&	0.501	&	0.246	&	0.094	&	0.695	&	0.763	&	0.392	&	0.232	&	0.149	&	0.599	&	0.654	&	0.350	&	0.195	&	0.086	&	0.623	&	0.674	\\
        \textbf{PiCANet}$_\text{CVPR18}$	&	0.536	&	0.200	&	0.084	&	0.769	&	0.749	&	0.356	&	0.166	&	0.155	&	0.609	&	0.584	&	0.322	&	0.173	&	0.083	&	0.649	&	0.643	\\
        \textbf{MSRCN}$_\text{CVPR19}$	&	0.443	&	0.074	&	0.091	&	0.637	&	0.686	&	0.454	&	0.128	&	0.133	&	0.618	&	0.669	&	0.419	&	0.101	&	0.073	&	0.641	&	0.706	\\
        \textbf{PFANet}$_\text{CVPR19}$	&	0.378	&	0.096	&	0.139	&	0.679	&	0.648	&	0.391	&	0.130	&	0.169	&	0.659	&	0.622	&	0.286	&	0.107	&	0.118	&	0.636	&	0.618	\\
        \textbf{HTC}$_\text{CVPR2019}$	&	0.204	&	0.071	&	0.129	&	0.517	&	0.489	&	0.174	&	0.076	&	0.172	&	0.477	&	0.442	&	0.221	&	0.099	&	0.088	&	0.548	&	0.520	\\
        \textbf{PoolNet}$_\text{CVPR2019}$	&	0.555	&	0.151	&	0.079	&	0.777	&	0.779	&	0.494	&	0.155	&	0.128	&	0.703	&	0.698	&	0.416	&	0.126	&	0.07	&	0.705	&	0.713	\\
        \textbf{ANet-SRM}$_\text{CVIU19}$	&	-	&	-	&	-	&	-	&	-	&	0.484	&	0.217	&	0.126	&	0.682	&	0.686	&	-	&	-	&	-	&	-	&	-	\\
        \textbf{CPD}$_\text{CVPR2019}$	&	0.706	&	0.383	&	0.052	&	0.853	&	0.868	&	0.550	&	0.306	&	0.115	&	0.726	&	0.730	&	0.508	&	0.286	&	0.059	&	0.747	&	0.771	\\
        \textbf{EGNet}$_\text{ICCV19}$	&	0.702	&	0.289	&	0.050	&	0.848	&	0.871	&	0.583	&	0.264	&	0.104	&	0.732	&	0.768	&	0.509	&	0.209	&	0.056	&	0.737	&	0.779	\\
        \textbf{SINet}$_\text{CVPR20}$	&	0.740	&	0.410	&	0.044	&	0.869	&	0.893	&	0.606	&	0.334	&	0.100	&	\textbf{0.752}	&	0.772	&	0.551	&	0.311	&	0.051	&	0.771	&	0.809	\\
        \hline	
        \textbf{BASNet}~(Ours)	&	\textbf{0.866} 	&	\textbf{0.650} 	&	\textbf{0.022}	&	\textbf{0.914}	&	\textbf{0.954}	&	\textbf{0.646}	&	\textbf{0.420}	&	\textbf{0.096}	&	0.749	&	\textbf{0.796}	&	\textbf{0.677}	&	\textbf{0.546}	&	\textbf{0.038}	&	\textbf{0.802}	&	\textbf{0.855}	\\
		\hline
	\end{tabular}
	\label{tab:compCOD}
\end{table*}

\subsubsection{Datasets}
We test our model on the CHAMELEON \cite{chameleon}, CAMO-Test \cite{le2019anabranch} and COD10K-Test datasets \cite{fan2020camouflaged}. CHAMELEON \cite{chameleon} contains 76 images taken by independent photographers. These images are marked as good examples of camouflaged animals by the photographers. 
CAMO \cite{le2019anabranch} contains both camouflaged and non-camouflaged subsets. We use the camouflaged subset, which comprises two further subsets: CAMO-Train (1,000 images) and CAMO-Test (250 images). 
COD10K \cite{fan2020camouflaged} is currently the largest camouflaged object detection dataset. It comprises 10,000 images of 78 object categories in various natural scenes. There are 5,066 images densely labeled with accurate (matting-level) binary masks. COD10K consists of 3,040 images for training (COD10K-Train) and 2,026 images for testing (COD10K-Test). For fair comparison, we use the same training sets as SINet \cite{fan2020camouflaged}. 



\subsubsection{Comparison with State-of-the-Arts}\label{sec:CmpCOD}
To validate the performance of the proposed BASNet on the camouflaged object segmentation task, we compared BASNet 
against 13 sate-of-the-art models, including FPN\cite{lin2017feature}, MaskRCNN\cite{he2017mask}, PSPNet\cite{zhao2017pyramid}, UNet++\cite{zhou2018unet++}, PiCANet\cite{liu2018picanet}, MSRCN\cite{huang2019mask}, PFANet\cite{zhao2019pyramid}, HTC\cite{chen2019hybrid}, PoolNet\cite{liu2019simple}, ANet-SRM\cite{le2019anabranch}, CPD\cite{wu2019cascaded}, EGNet\cite{zhao2019egnet} and SINet\cite{fan2020camouflaged}. For fair comparison, the results of different models are either provided by the authors or obtained by re-training the model with the default settings with same training data. 

\begin{figure}[t!]
	\centering
	\begin{overpic}[width=.9\columnwidth]{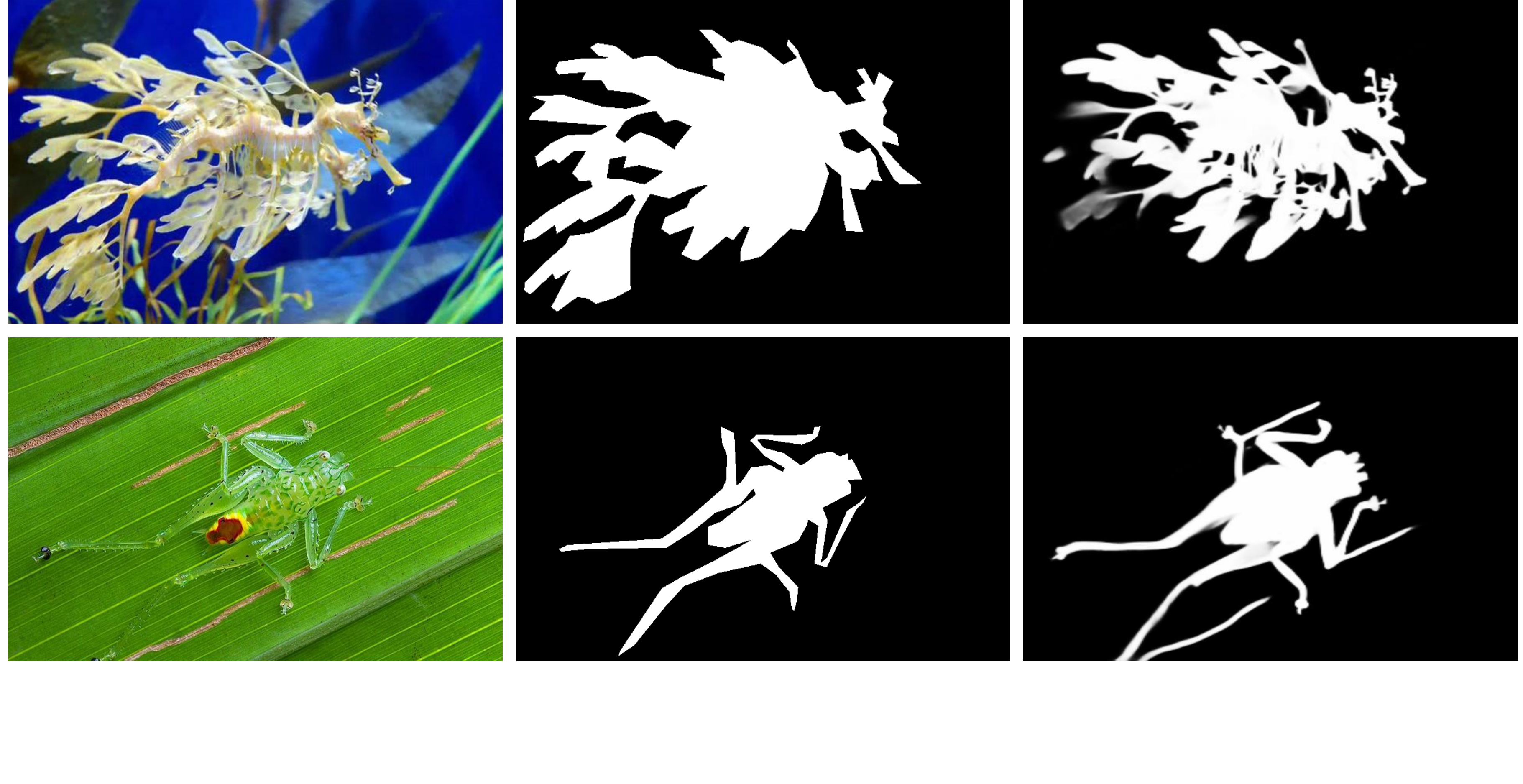}
	\put(9,2){\small (a) Image}
	\put(45,2){\small (b) GT}
	\put(75,2){\small (c) BASNet}
  \end{overpic}
  \vspace{-5pt}
  \caption{Failure cases on camouflaged object segmentation task. The first row shows the typical false negative artifacts. The second row illustrates the false positive phenomenon.
  }\label{fig:fail_cod}
\end{figure}

\textbf{Quantitative Evaluation:} 
The quantitative evaluation results are illustrated in Table \ref{tab:compCOD}. As we can see that our BASNet achieves the best performance in nearly all metrics with great advantages. SINet is the second best model. EGNet and CPD are competitive with each other and can be ranked as the third and fourth best. Our BASNet improves the weighted F-measure $F^w_\beta\uparrow$ with large margins ($12.6\%$, $4.0\%$ and $12.6\%$ on CHAMELEON, CAMO-Test and COD10K-Test respectively). Particularly, our BASNet outperforms the second best model SINet by $24.0\%$, $8.6\%$ and $23.5\%$ in terms of the relax boundary F-measure $F^b_\beta\uparrow$ on CHAMELEON, CAMO-Test and COD10K-Test datasets. This reveals the effectiveness and accuracy of our BASNet in capturing boundaries and fine structures. In terms of the $M\downarrow$, our BASNet reduces the metric by $50.0\%$, $4.0\%$ and $25.5\%$ on the three datasets, respectively. For the structure measures, $S_{\alpha}\uparrow$, the improvements of our BASNet against the second best model are also significant ($4.5\%$ and $3.1\%$ on CHAMELEON and COD10K-Test datasets) but there is a $0.3\%$ $S_{\alpha}$ decrease on CAMO-Test. Compared with SINet, $E^m_{\phi}\uparrow$ takes both local and global structure similarity into consideration. As we can see, our BASNet achieves even greater improvements ($6.1\%$, $2.4\%$ and $4.6\%$ on the three datasets, respectively) against the second best model in $E^m_{\phi}\uparrow$ than in $S_{\alpha}\uparrow$.

\begin{figure*}[t!]
	\centering
	\begin{overpic}[width=.9\textwidth]{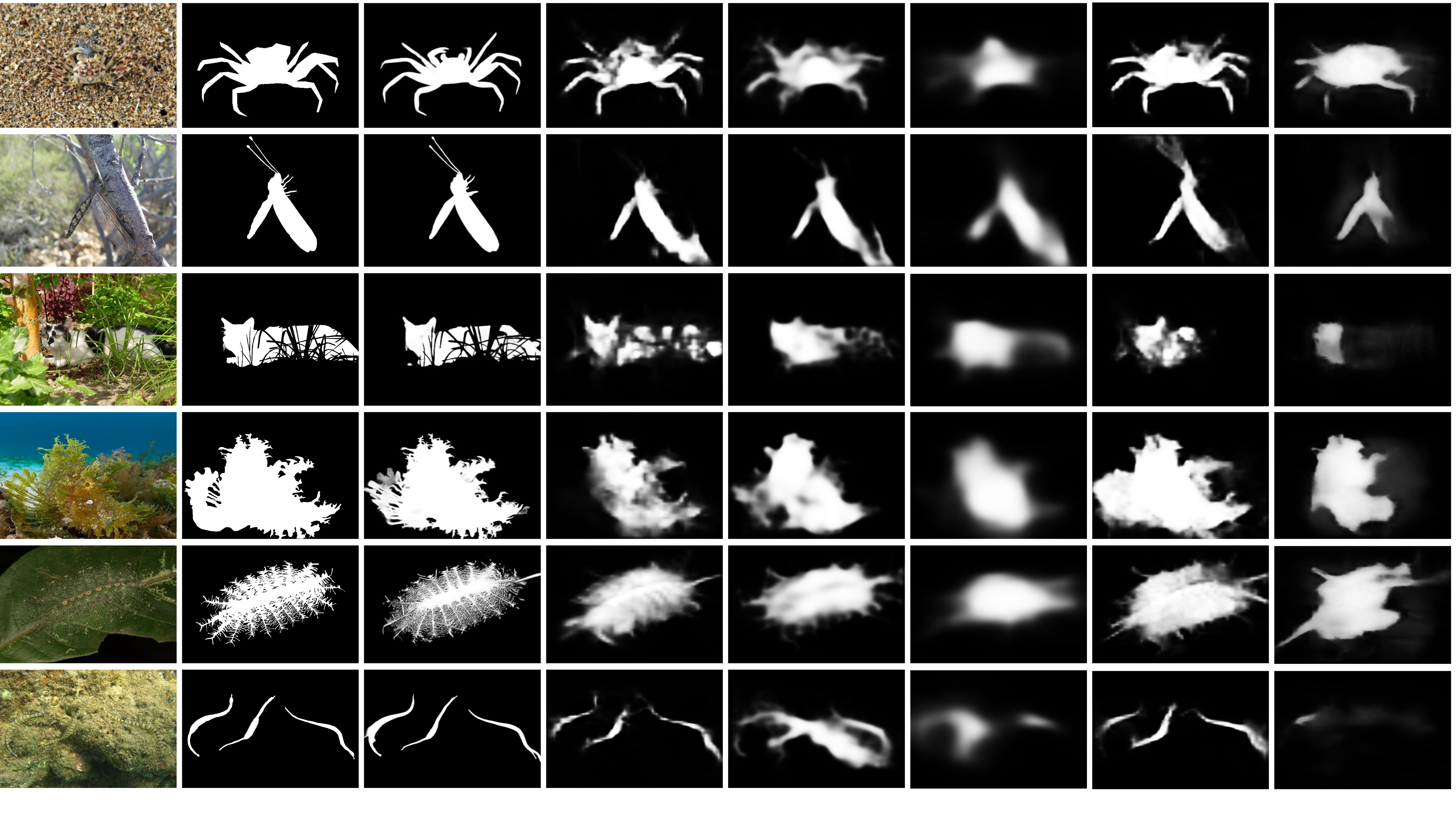}
	\put(3,1){\small (a) Image}
	\put(16,1){\small (b) GT}
	\put(27,1){\small (c) BASNet}
	\put(39.5,1){\small (d) SINet}
	\put(52,1){\small (e) EGNet}
	\put(64,1){\small (f) PoolNet}
	\put(78,1){\small (g) CPD}
	\put(89,1){\small (h) PiCANet}
  \end{overpic}
  \caption{Qualitative comparison on camouflaged object segmentation datasets. See \secref{sec:CmpCOD} for details.
  }\label{fig:qual-cod}
\end{figure*}
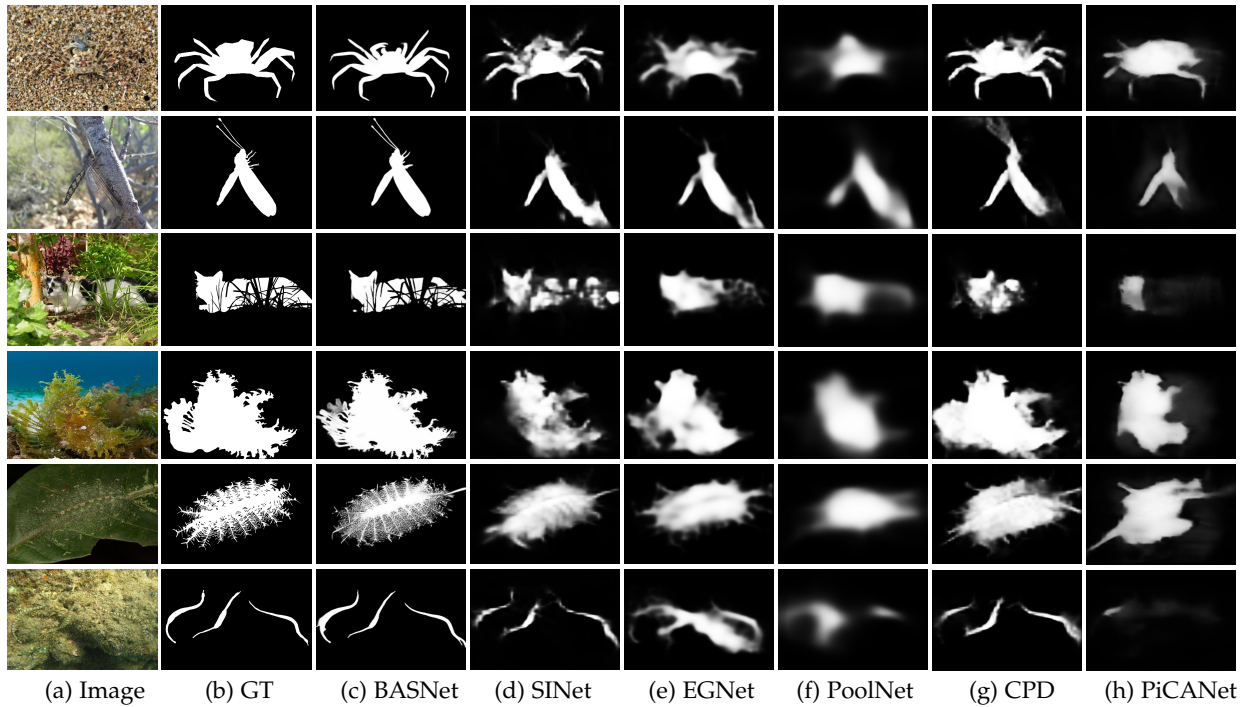

\textbf{Qualitative Evaluation:} 
Qualitative comparisons against several of the baseline models are illustrated in Fig. \ref{fig:qual-cod}. As we can see, our BASNet (the 3rd column) is able to handle different types of challenging camouflaged cases including complex foreground targets with low contrast (the 1st row), targets with very thin foreground structures (the 2nd and 5th row), targets occluded by fine objects (the 3rd row), targets with complicated boundaries (the 4th row), targets with extremely complex hollow structures (the 5th row), multiple objects with low contrast (the 6th row), \etc. 
Compared with the results of other models, the results of our BASNet demonstrates its excellent abilities of perceiving fine structures and complicated boundaries, which also explains why our BASNet is able to achieve such high boundary evaluation scores $F^b_\beta\uparrow$ on camouflage object segmentation datasets (see \tabref{tab:compCOD}). 

\subsubsection{Failure Cases}
Although our BASNet outperforms other camouflaged object segmentation (COS) models and rarely produces completely incorrect results, there are still some false negative (the 1st row in Figure \ref{fig:fail_cod}) and false positive predictions (the 2nd row in Fig. \ref{fig:fail_cod}) in many of the COS cases. It is worth noting that other models usually have the same or even worse results on these challenging cases. Although these failure cases may not have a huge impact on evaluation metrics, they will somehow limit the applications and degrade the user experiences.

\section{Applications}
Thanks to the high accuracy, fast speed and simple architecture of our network, we developed two real-world applications based on BASNet: \textbf{AR COPY \& PASTE} and \textbf{OBJECT CUT}. These two applications further demonstrate the effectiveness and efficiency of our BASNet.

\subsection{Application I: AR COPY \& PASTE}
\begin{figure}[t!]
	\centering
	\begin{overpic}[width=.9\columnwidth]{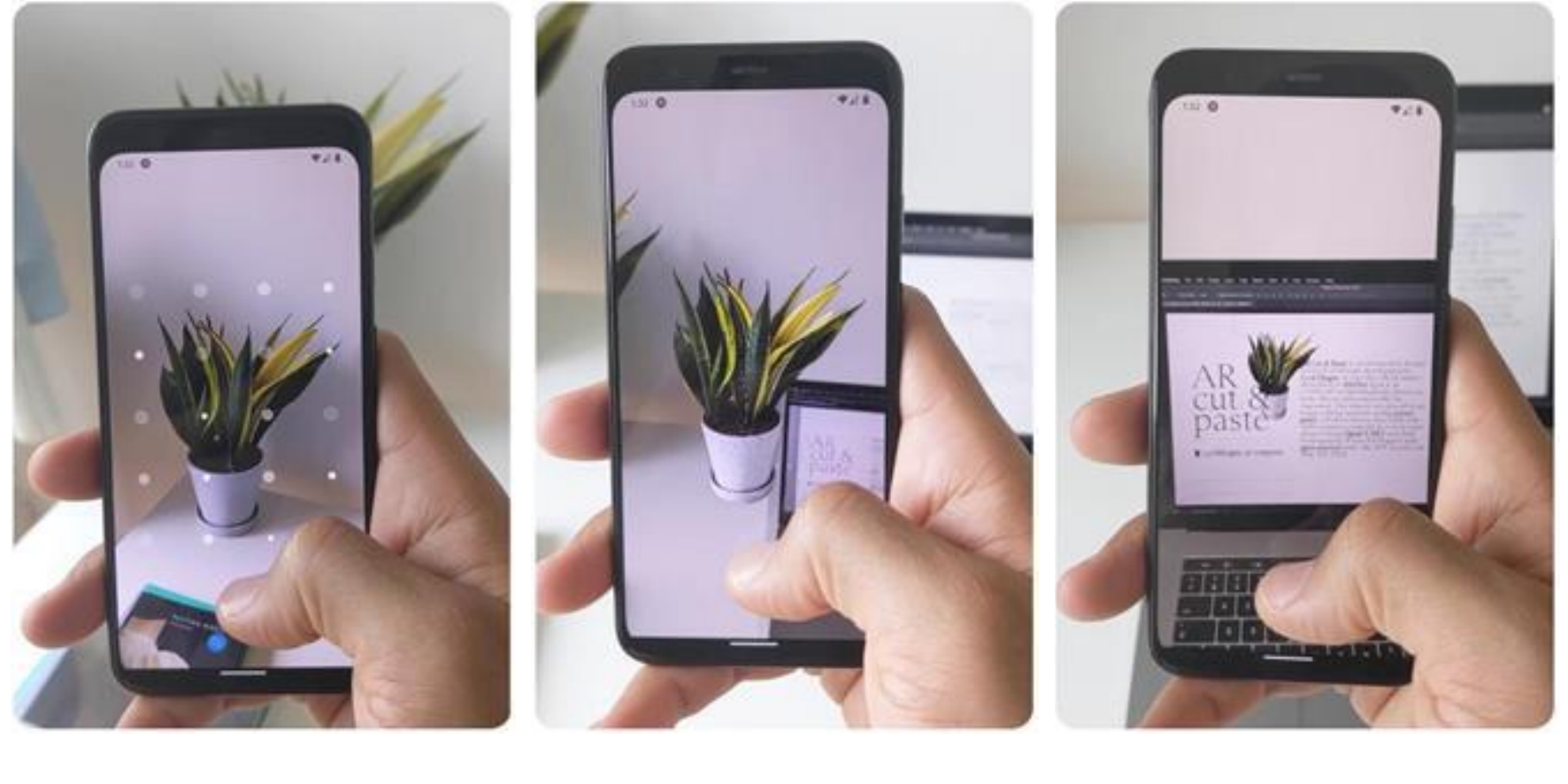}
	\put(9,-1){\small (a) Copy}
	\put(45,-1){\small (b) Move}
	\put(76,-1){\small (c) Paste}
  \end{overpic}
  \vspace{0.5pt}
  \caption{Screenshots from the video demonstration. (a) \textbf{Copy}: Point and tap to ``copy'' the object by masking its background out using BASNet. (b) \textbf{Move}: Move the mobile phone, where the ``copied'' object is shown, to target at the computer screen at a specific location. (c) \textbf{Paste}: Tap to ``paste'' the ``copied'' object into the current document.
  }\label{fig:arcopypaste_steps}
\end{figure}

Introduced in HCI by Larry Tesler in the early 70s \cite{tesler2012personal}, cut/copy-paste has become essential to many applications of modern computing. In this section, we explore how BASNet can be used to apply this principle to the mobile phone camera and seamlessly transfer visual elements between the physical space and the digital space. AR COPY \& PASTE is a prototype that we built upon our BASNet to conveniently capture real-world items using the camera of a mobile phone (\eg~objects, people, drawings, schematics, \etc), automatically remove the background, and insert the result in an editing software on the computer by pointing the camera at it, as shown in Fig. \ref{fig:arcopypaste_steps}. Specifically, AR COPY \& PASTE first removes the background of the photo and only shows the foreground salient target on the mobile phone screen. Then users can ``paste'' the segmented target by moving the cellphone to point the mobile camera at a specific location of a document opened on a computer. The whole process of AR COPY \& PASTE makes it seem like the real-world target is ``copied'' and ``pasted'' into a document, which provides a novel and inspiring interactive experience. A demonstration video\footnote{\url{https://twitter.com/cyrildiagne/status/1256916982764646402}} of the prototype has been released along with the source code.\footnote{\url{https://github.com/cyrildiagne/ar-cutpaste}} Both have received world-wide attention (millions of views for the video, tens of thousands of Github stars for the source code). Hundreds of thousands of people have subscribed to the private beta. 


\subsubsection{Workflow of AR COPY \& PASTE}

\begin{table*}[t!]
    \centering
    \renewcommand{\arraystretch}{0.5}
    \setlength\tabcolsep{18pt}
    \caption{Number of operations of different methods for image capturing and masking out.}
    \begin{tabular}{c|cc}
        \hline 
        Methods & Number of steps (on mobile) & Number of steps (on desktop) \\
        \hline 
        \tabincell{c}{Cross-platform \\(Android v11 and macOS v10.15)} & \tabincell{l}{6\\ 
        \textcircled{1} Take a photo, \\ \textcircled{2} Tap the thumbnail, \\ \textcircled{3} Tap share, \\ \textcircled{4} Tap more, \\ \textcircled{5} Select Bluetooth, \\ \textcircled{6} Tap the destination on a device.} & \tabincell{l}{ 5 \\ 
        \textcircled{1} Click ``Open'' on the Bluetooth notification, \\
        \textcircled{2} Export image, \\
        \textcircled{3} Select destination software, \\
        \textcircled{4} Use ``Select Object'' Tool, \\
        \textcircled{5} Apply mask.} \\
        \hline
        \tabincell{c}{Constructor-specific \\ (iOS v13 and macOS v10.15)} & 
        \tabincell{l}{5 \\ 
        \textcircled{1} Take a photo, \\
        \textcircled{2} Tap the thumbnail, \\
        \textcircled{3} Tap share, \\
        \textcircled{4} Tap “AirDrop”, \\ \textcircled{5} Tap the destination on a device.} &
        \tabincell{l}{5 \\ 
        \textcircled{1} Click “Open” on the AirDrop notification, \\ 
        \textcircled{2} Export image, \\ \textcircled{3} Select destination software, \\
        \textcircled{4} Use ``Select Object'' Tool, \\
        \textcircled{5} Apply mask.} \\
        \hline 
        AR COPY \& PASTE (Ours) & \tabincell{l}{\hspace{-0.2in}2 \\
        \hspace{-0.2in}\textcircled{1} Take a photo, \\
        \hspace{-0.2in}\textcircled{2} Tap toward the destination \\
        software.} & \tabincell{l}{0 \\
        } \\
        \hline 
    \end{tabular}
    \label{tab:arcopypaste_comp}
\end{table*}

\begin{figure}[t!]
    \begin{center}
        \includegraphics[width=\linewidth]{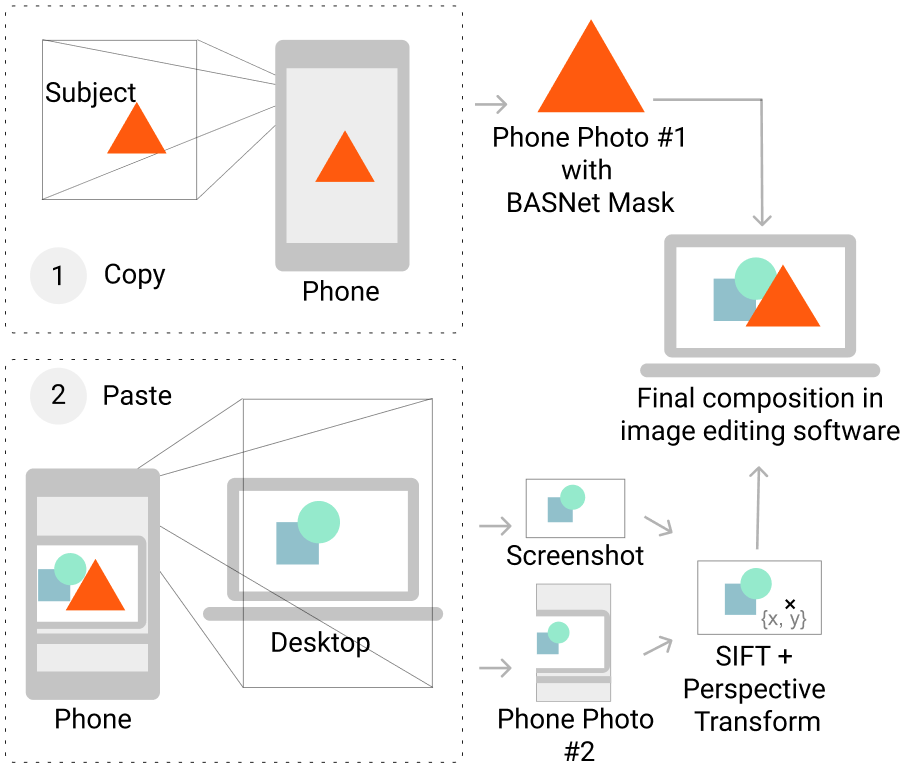}
    \end{center}
    \caption{Schematic of the AR COPY \& PASTE flow.}
    \label{fig:arcopypaste}
\end{figure}

From the perspective of users, our AR COPY \& PASTE only consists of two main steps: ``copy'' and ``paste'', as shown in Fig. \ref{fig:arcopypaste}. \textbf{(1) Copy.} The first step consists in pointing the mobile camera at a subject and tapping the screen to take a picture. BASNet is then used to hide all the pixels that are not part of the main foreground subject. The remaining pixels keep floating on top of the camera and provide a preview of the paste result. Compared to other methods like image segmentation \cite{minaee2020image}, BASNet can produce very accurate segmentation results with sharp and high-quality boundaries, which is essential in many image composition workflows. \textbf{(2) Paste.} The second step consists of pointing the mobile phone at a specific location on the computer screen and tapping to ``paste'' the ``copied'' subject. SIFT \cite{lowe2004distinctive} (implemented in OpenCV \cite{opencv_library}) is used to find the corresponding computer screen coordinates targeted by the center of the mobile phone camera. The image containing the background removed target is finally imported in an image editing software at the computed screen coordinates to create the final composition.


\begin{figure}[thbp]
    \begin{center}
        \includegraphics[width=\linewidth]{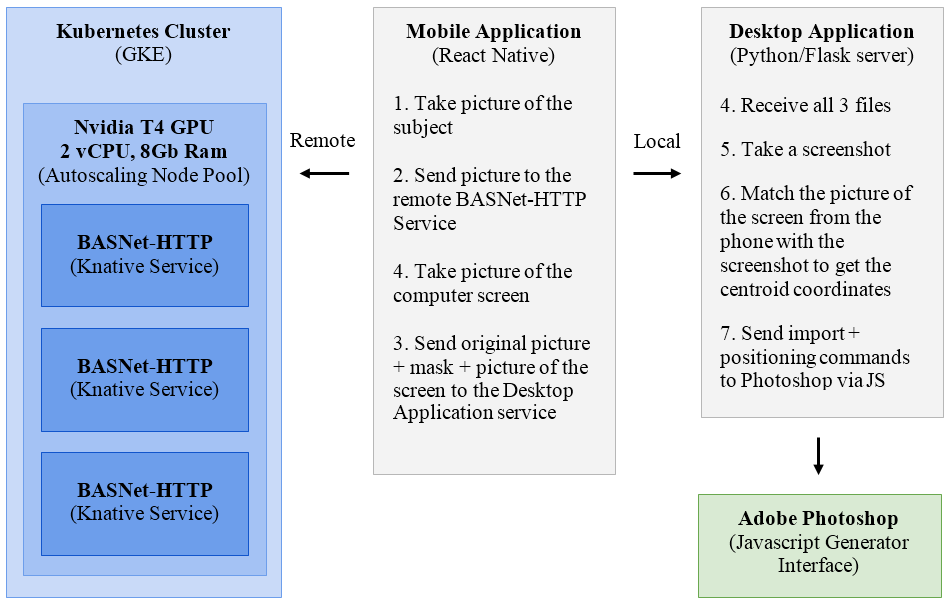}
    \end{center}
    \caption{Overall pipeline of the AR COPY \& PASTE.}
    \label{fig:arcopypaste_pip}
\end{figure}

\begin{figure}[t!]
	\centering
	\begin{overpic}[width=\columnwidth]{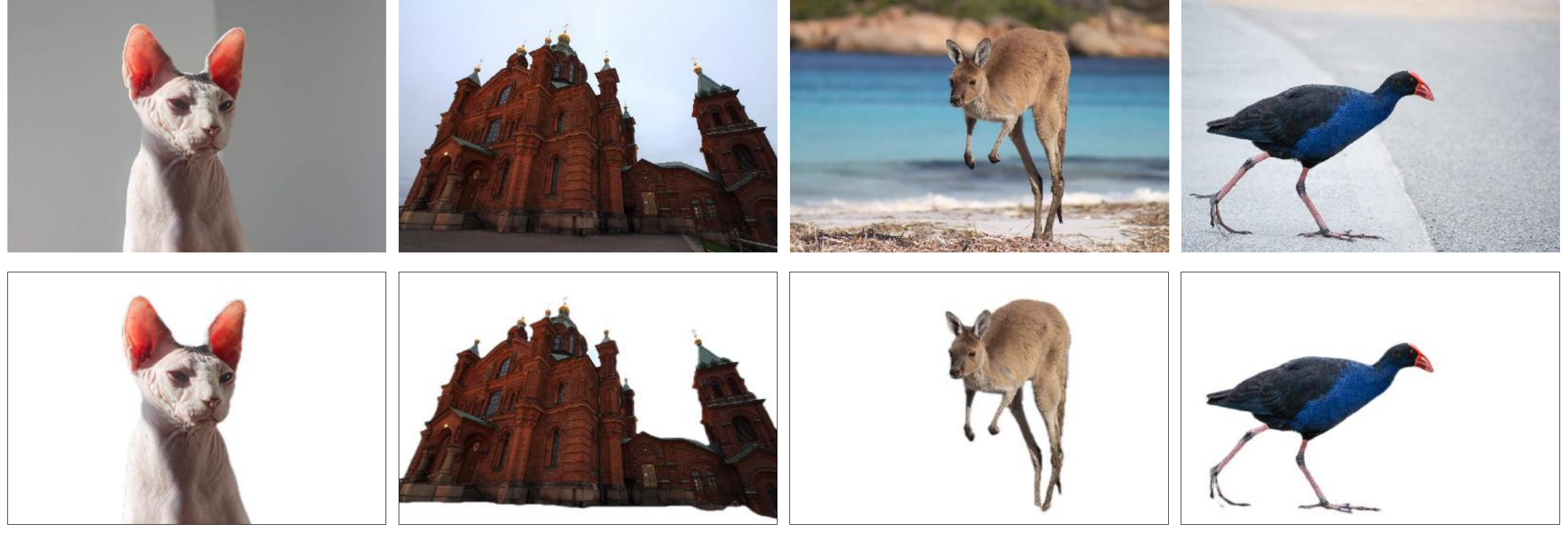}
  \end{overpic}
  \vspace{-1pt}
  \caption{Sample results given by the OBJECT CUT API: the first row shows the input images and the second row shows the background removed results.
  }\label{fig:object_cut_results}
\end{figure}


\subsubsection{Implementation Details} 
Fig. \ref{fig:arcopypaste_pip} illustrates the overall implementation pipeline of the AR COPY \& PASTE prototype, which consists of three main parts: Kubernetes cluster, mobile application and desktop application. The AR COPY \& PASTE prototype was built using our BASNet model trained on DUTS-TR \cite{wang2017learning}. To make sure that it runs smoothly on mobile device, it has been wrapped as an HTTP service/container image\footnote{\url{https://github.com/cyrildiagne/basnet-http}} so that it can be deployed easily on remote GPUs using Kubernetes \footnote{\url{https://github.com/kubernetes/kubernetes}}. Hence, photos taken by mobile devices in AR COPY \& PASTE are sent to the remote server to obtain their segmentation masks. The desktop application is a python HTTP server which takes three files from the mobile application as input (original picture, BASNet mask, photo of the screen) and runs SIFT feature matching and perspective transformation based on the photo of the screen and the screenshot to determine the destination coordinates. Finally, the desktop application is responsible for sending javascript commands to desktop apps like Photoshop\footnote{\url{https://www.photoshop.com/en}} in order to import the image into the document at the right position.

\begin{figure}[t!]
    \begin{center}
        \includegraphics[width=\linewidth]{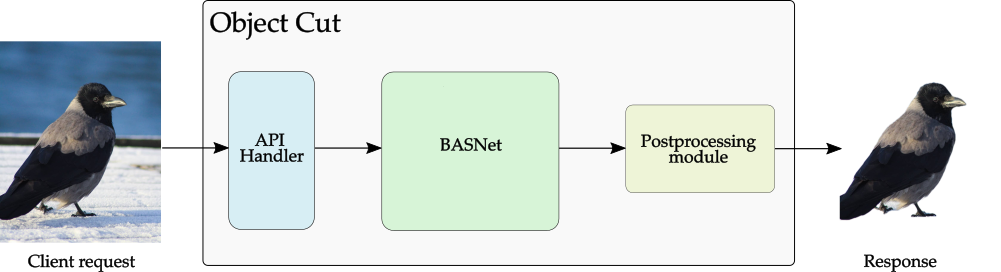}
    \end{center}
    \caption{OBJECT CUT pipeline.}
    \label{fig:object_cut}
\end{figure}

\subsubsection{Comparison with Other Methods}
Our AR COPY \& PASTE prototype applies BASNet in a novel human-computer-interaction setting, which makes the process easier and faster than other methods (two operations for our method versus 10 or 11 operations for other methods). Table \ref{tab:arcopypaste_comp} illustrates examples of typical user flow to clip and import an object from a mobile devices to an desktop image editing software, such as Adobe Photoshop. As we can see, our prototype greatly reduces the numbers of operations and simplifies the process. Besides, our AR COPY \& PASTE allows users to delegate some of the lower-level decisions (how visible each pixel should be), and focus on the higher-level objectives (how do they want the object to look). Removing these tasks lowers the barrier to entry (there is no need to learn how to paint masks in an image editing software), saves a significant amount of time, and ultimately leads to better end results by removing the cognitive load of the low-level tasks \cite{carter2017using}. 






\subsection{Application II: OBJECT CUT}
OBJECT CUT is an online image background removal service that uses BASNet. Removing the background from an image is a common operation in the daily work of professional photographers and image editors. This process is usually a repeatable and manual task that requires a lot of effort and time. However, thanks to BASNet, one of the most robust and fastest performing deep learning models for image segmentation, OBJECT CUT was able to turn it into an easy and automatic process. 
The program was built as an API to make it as easy as possible to integrate. APIs, also known as Application Programming Interfaces, are already commonly used to integrate different types of solutions to improve systems without actually knowing what is happening inside. For instance, RESTful APIs are a standard in the software engineering field for designing and specifying APIs. Making it substantially easier to adapt desired APIs to specific workflows.

Our system is based on three well-distinguished parts, as shown in Fig. \ref{fig:object_cut}: 
1) the API Handler, responsible for receiving and validating clients requests, downloading and preprocessing input images and sending those to the model; 
2) BASNet, responsible for performing salient object detection. 
3) Once the output from the network is generated, the postprocessing module applies an unsharp masking algorithm and morphological operations to improve the quality of the output. Afterward, OBJECT CUT uploads the cropped image to the Google Cloud Storage and returns its public URL to the client. This is structured as-is in order to isolate different parts, such as the BASNet component, removing all the API parameter validations as well as image download and upload processes, as much as possible. In this scenario, OBJECT CUT maximizes the operations running on the BASNet thread. The whole stack from the API is running under Docker containers, all managed by the cloud native application proxy called Traefik. The usage of Traefik here allows us to have a production-ready deployment making easy, from the containers' perspective, to communicate with other processes. In addition, we have a Load Balancer system in place to enable each of the components to scale more easily. The source code for this pipeline can be found under the OBJECT CUT GitHub repository: \url{https://github.com/AlbertSuarez/object-cut}. 

To ensure easy integration, it is publicly available at RapidAPI\footnote{https://rapidapi.com/objectcut.api/api/background-removal}, the biggest API marketplace in the world, and it has been effectively utilized by people and companies from 86 different countries around the globe, including China, United States, Canada, India, Spain, Russia, \etc. OBJECT CUT was born to power up the designing and image editing process for the people who work with images daily. Integrating the OBJECT CUT API removes the necessity of understanding the complex inner workings behind it and automates the process of removing the background from images in a matter of seconds. See examples in Fig. \ref{fig:object_cut_results}.


\section{Conclusion}
In this paper, we proposed a novel end-to-end boundary-aware model, BASNet, and a hybrid fusion loss for accurate image segmentation. 
The proposed BASNet is a predict-refine architecture, which consists of two components: a prediction network and a refinement module. 
Combined with the hybrid loss, BASNet is able to capture both large-scale and fine structures, \eg~thin regions, holes, and produce segmentation probability maps with highly accurate boundaries. 
Experimental results on six salient object segmentation datasets, one salient object in clutter dataset and three camouflaged object segmentation datasets demonstrate that our model achieves very competitive performance in terms of both region-based and boundary-aware measures against other models.  
Additionally, the proposed network architecture is modular. It can be easily extended or adapted to other tasks by replacing either the prediction network or the refinement module. 
The two (close to) commercial applications, AR COPY \& PASTE and OBJECT CUT, based on our BASNet not only prove the effectiveness and efficiency of our model but also provide two practical tools for reducing the workload in real-world production scenarios. The world-wide impacts of these two applications indicates the huge demand for highly accurate segmentation approaches, which motivates us to explore more accurate and reliable segmentation models.  




\ifCLASSOPTIONcaptionsoff
  \newpage
\fi

\bibliographystyle{ieee_fullname}
\bibliography{xuebin}

\vspace{-30pt}
\begin{IEEEbiographynophoto}{Xuebin QIN} obtained his PhD degree from the University of Alberta, Edmonton, Canada, in 2020. Since March, 2020, He is a Postdoctoral Fellow in the Department of Computing Science and the Department of Radiology and Diagnostic Imaging, University of Alberta, Canada. His research interests include highly accurate image segmentation, salient object detection, image labeling and detection. He has published about 10 papers in vision and robotics conferences such as CVPR, BMVC, ICPR, WACV, IROS, etc.
\end{IEEEbiographynophoto}

\vspace{-30pt}
\begin{IEEEbiographynophoto}
{Deng-Ping FAN} received his PhD degree from the Nankai University in 2019.
He joined Inception Institute of Artificial Intelligence (IIAI) in 2019.
He has published about 25 top journal and conference papers such as CVPR, ICCV, ECCV, etc. 
His research interests include computer vision and visual attention, especially on 
RGB salient object detection (SOD), RGB-D SOD, Video SOD, Co-SOD. 
He won the Best Paper Finalist Award at IEEE CVPR 2019, 
the Best Paper Award Nominee at IEEE CVPR 2020.
\end{IEEEbiographynophoto}

\vspace{-30pt}
\begin{IEEEbiographynophoto}{Chenyang Huang} obtained his M.Sc. degree from the University of Alberta, Edmonton, Canada, in 2019. He is currently pursuing a Ph.D. degree in the Department of Computing Science of the same university. His research is mainly focusing on deep learning, natural language processing, and computer vision. He has publications on some prestigious conferences such as NAACL and CVPR.
\end{IEEEbiographynophoto}

\vspace{-30pt}
\begin{IEEEbiographynophoto}{Cyril Diagne} is a designer and coder and a co-founder of Init ML, a company that brings machine learning to production through practical uses, such as ClipDrop.
Cyril is a former Professor and Head of Media \& Interaction Design at ECAL (Lausanne University of Arts \& Design, Switzerland) where he continues to give regular workshops.
In 2015, he started a residency at Google Arts \& Culture, where he helped kickstart the Google Arts Experiments initiative and created multiple machine learning projects such as the viral phenomenon Art Selfie.
\end{IEEEbiographynophoto}

\vspace{-30pt}
\begin{IEEEbiographynophoto}{Zichen Zhang} is a Ph.D. student in Statistical Machine Learning at the University of Alberta. He obtained his M.Sc degrees from Dalhousie University and the University of Alberta and B.E degree from Huazhong University of Science and Technology.  He’s interested in machine learning and its applications in robotics perception and control.
\end{IEEEbiographynophoto}

\vspace{-30pt}
\begin{IEEEbiographynophoto}{Adrià Cabeza Sant'Anna} is a computer engineer who graduated at Universitat Politècnica de Catalunya, BarcelonaTech in Computer Science. His current position is Deep Learning Engineer at restb.ai, a Computer Vision company for Real Estate based in Barcelona. Previously, he worked as an Algorithmic Methods of Data Mining grader assistant in the Department of Computer Science at Aalto University, Helsinki. He was the president of the Student Representatives Association at Barcelona School of Informatics. His research interests include machine learning, computer vision, generative models, and data mining. He has co-developed ObjectCut and participates in the organization of HackUPC, the biggest student-run hackathon in Europe, located at Barcelona School of Informatics.
\end{IEEEbiographynophoto}

\vspace{-30pt}
\begin{IEEEbiographynophoto}{Albert Suàrez} is a software engineer who graduated at Universitat Politècnica de Catalunya, BarcelonaTech in Software Engineering. His current position is Principal Software Engineer at restb.ai, a Computer Vision company for Real Estate based in Barcelona, Spain. He was the co-director of the biggest student-run hackathon in Europe, called HackUPC, located at Barcelona School of Informatics.
\end{IEEEbiographynophoto}

\vspace{-30pt}
\begin{IEEEbiographynophoto}
{Martin Jagersand's} research interests are in Robotics, Computer Vision, and Graphics, especially vision guided motion control and vision-based human-robot interfaces. He studied physics at Chalmers Sweden (MSc 1991). He was awarded a Fulbright fellowship for graduate studies in the USA. He studied Computer Science at the Univ. of Rochester, NY (MSc 1994, PhD 1997). He held an NSF CISE postdoc fellowship, at Yale University, and then was a research faculty in the Engineering Research Center for Surgical Systems and Technology at Johns Hopkins University. He is now a faculty member at the University of Alberta.
\end{IEEEbiographynophoto}

\vspace{-30pt}
\begin{IEEEbiographynophoto}
{Ling Shao} 
is currently the CEO and the Chief
Scientist of the Inception Institute of Artificial
Intelligence, Abu Dhabi, United Arab Emirates.
He is also the Executive Vice President and a
Provost of the Mohamed bin Zayed University of
Artificial Intelligence. His current research interests include computer vision, machine learning,
and medical imaging. Dr. Shao is a fellow of
IAPR, IET, and BCS. He is an Associate Editor of the IEEE TRANSACTIONS ON IMAGE PROCESSING, the IEEE TRANSACTIONS ON
NEURAL NETWORKS AND LEARNING SYSTEMS, and several other
top journals.
\end{IEEEbiographynophoto}

\vfill
\end{document}